\documentclass[10pt,twocolumn,letterpaper]{article}

\usepackage[pagenumbers]{cvpr}

\usepackage{graphicx}
\usepackage{amsmath}
\usepackage{amssymb}
\usepackage{booktabs}
\usepackage{multirow}
\usepackage{makecell}
\usepackage{float}
\usepackage{diagbox}

\usepackage{graphics}
\usepackage{sidecap}
\usepackage{setspace}
\usepackage[linesnumbered,ruled,vlined]{algorithm2e}

\usepackage[pagebackref,breaklinks,colorlinks]{hyperref}
\usepackage{paralist}

\usepackage[capitalize]{cleveref}
\crefname{section}{Sec.}{Secs.}
\Crefname{section}{Section}{Sections}
\Crefname{table}{Table}{Tables}
\crefname{table}{Tab.}{Tabs.}
\crefname{line}{Algo.}{lines}

\usepackage[accsupp]{axessibility} %

\begin{document}

\title{LayoutDiffusion:  Improving Graphic Layout Generation by \\ Discrete Diffusion Probabilistic Models}

\author{
Junyi Zhang$^1$\thanks{Work done during an internship at Microsoft Research Asia.} \qquad Jiaqi Guo$^2$ \qquad Shizhao Sun$^2$ \qquad Jian-Guang Lou$^2$ \qquad Dongmei Zhang$^2$\\
$^1$Shanghai Jiao Tong University \qquad $^2$Microsoft Research Asia\\
{\tt\small junyizhang@sjtu.edu.cn \qquad \{jiaqiguo,~shizsu,~jlou,~dongmeiz\}@microsoft.com}
}
\maketitle

\begin{abstract}
    Creating graphic layouts is a fundamental step in graphic designs.
    In this work, we present a novel generative model named LayoutDiffusion for automatic layout generation.
    As layout is typically represented as a sequence of discrete tokens, LayoutDiffusion models layout generation as a discrete denoising diffusion process.
    It learns to reverse a mild forward process, in which layouts become increasingly chaotic with the growth of forward steps and layouts in the neighboring steps do not differ too much.
    Designing such a mild forward process is however very challenging as layout has both categorical attributes and ordinal attributes.
    To tackle the challenge, we summarize three critical factors for achieving a mild forward process for the layout, i.e., legality, coordinate proximity and type disruption.
    Based on the factors, we propose a block-wise transition matrix coupled with a piece-wise linear noise schedule.
    Experiments on RICO and PubLayNet datasets show that LayoutDiffusion outperforms state-of-the-art approaches significantly. 
    Moreover, it enables two conditional layout generation tasks in a plug-and-play manner without re-training and achieves better performance than existing methods.
    Project page: \url{https://layoutdiffusion.github.io}.
\end{abstract}
\vspace{-5px}
\section{Introduction}
\label{sec:intro}

\begin{figure}[t]
  \centering
  \begin{subfigure}{1\linewidth}
   \includegraphics[width=1\linewidth]{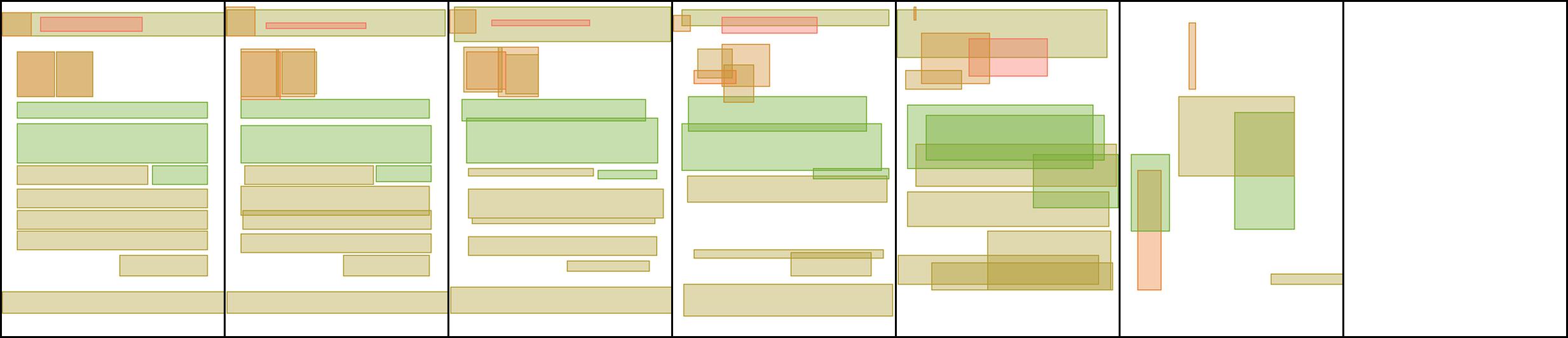}
    \caption{Mild forward corruption process}
    \label{noise_ours}
  \end{subfigure}
  \hfill
  \begin{subfigure}{1\linewidth}
   \includegraphics[width=1\linewidth]{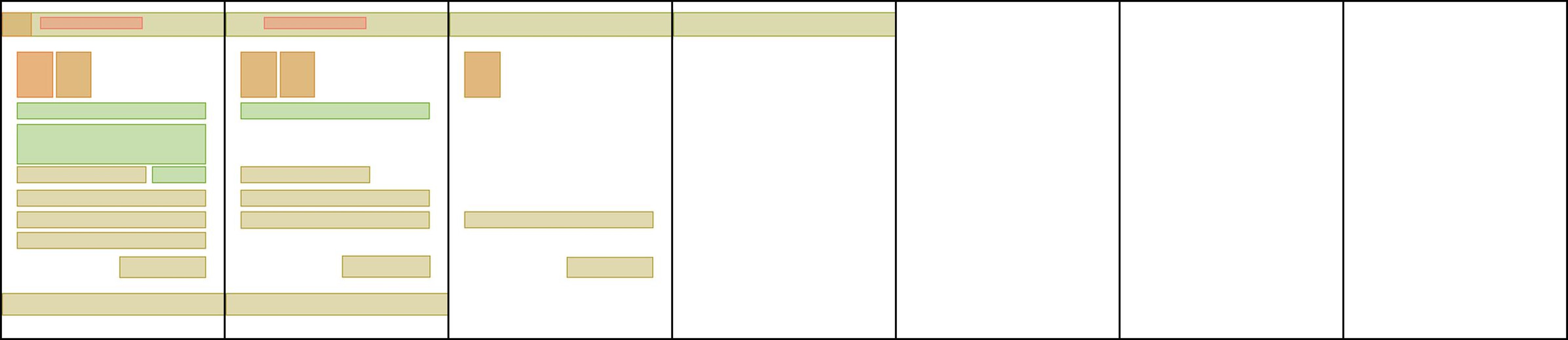}
    \caption{Absorbing~\cite{d3pm}}
    \label{noise_absorb}
  \end{subfigure}
  \hfill
  \begin{subfigure}{1\linewidth}
   \includegraphics[width=1\linewidth]{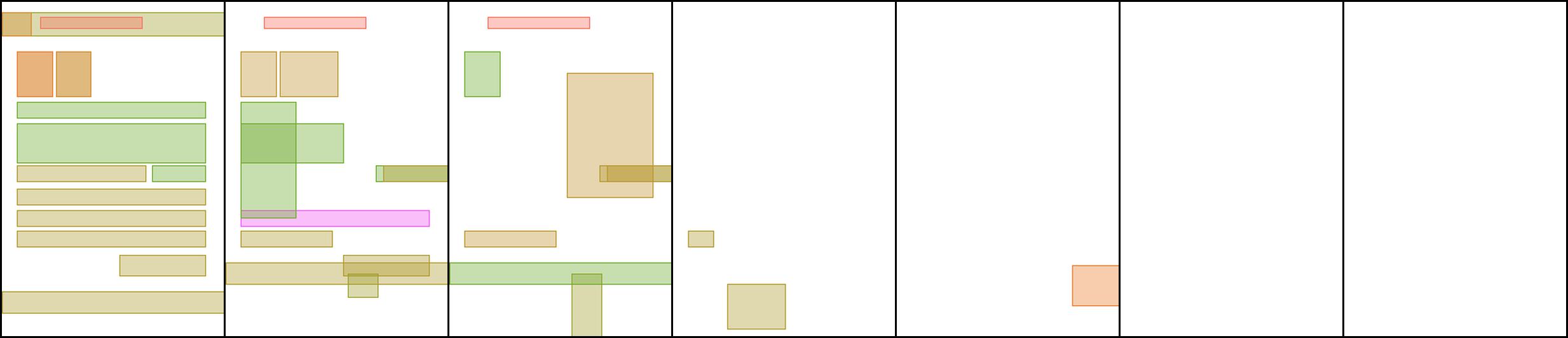}
    \caption{Uniform~\cite{2015}}
    \label{noise_uniform}
    \hfill
        \begin{subfigure}{1\linewidth}
    \centering
   \includegraphics[width=1\linewidth]{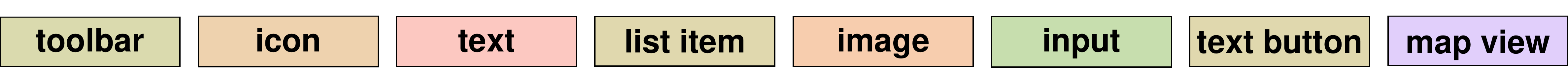}
    \caption{The mapping between colors and element types used in \cref{noise_ours,noise_absorb,noise_uniform}}
    \label{graphic layout eg}
  \end{subfigure}
  \end{subfigure}
 \vspace{-15px}
\caption{Comparison of different forward corruption processes. We sample the layouts at the timesteps 0, 1/6, 2/6, 3/6, 4/6, 5/6, and 1 of the total timestep. The blank page is used when the format of the layout sequence is destroyed.} %
   \label{noise}
   \vspace{-12.5px}
\end{figure}

Graphic layout, i.e., the \emph{sizes and positions} of elements, is important to the interaction between the viewer and the information.
Recently, layout generation attracts growing research interest.
Leading approaches~\cite{layouttransformer,unilayout,blt,c2f} often represent a layout as a sequence of elements and leverage Transformer~\cite{vaswani2017attention} to model element relationships.
As the placement of one element could depends on any part of a layout, \emph{global context modeling} plays a critical role in layout generation. 
However, there is no satisfactory solution to it.
Some studies simply consider biased context~\cite{layouttransformer,unilayout,vtn,c2f}.
They generate layout sequences autoregressively, where the generation order for elements is predefined and the placement of one element only depends on a certain part of layout.
A few other studies try to utilize global context by non-autoregressive generation~\cite{ndn}.
Unfortunately, they fail to improve the generation quality significantly since it is too challenging to generate a sequence in a single pass~\cite{ghazvininejad2019mask}.

Meanwhile, the emerging diffusion probabilistic model (DDPM)~\cite{ddpm,song2020score} achieves amazing performance on many generation tasks~\cite{amazingdiffusion1,amazingdiffusion2,amazingdiffusion3,amazingdiffusion4,amazingdiffusion5,amazingdiffusion6,amazingdiffusion7}.
It consists of multiple rounds, each of which gradually denoises the latent variables towards the desired data distribution.
This sort of process seems to be a promising solution to layout generation.
First, the layout generated in the last round could serve as the global context for the generation in the next round.
Second, by multiple rounds of denoising, a layout could be refined iteratively, overcoming the challenge of single-pass generation from non-autoregressive models.

To this end, we propose \emph{LayoutDiffusion} to improve graphic layout generation.
As a layout is represented as a sequence of discrete tokens~\cite{unilayout,blt,layouttransformer}, we formulate layout generation as a discrete diffusion process.
Roughly speaking, it samples a layout by reversing a forward process.
The forward process corrupts the real data into a sequence of increasingly noisy latent variables by a fixed Markov Chain.
The reverse process starts from noise and denoises it step by step via learning the posterior distribution.

To ease the estimation of the posterior distribution, it is critical to design a \emph{mild} forward corruption process~\cite{improvedDdpm}, in which latent variables in neighboring steps do not differ too much and become increasingly chaotic with the growth of forward steps (see~\cref{noise_ours}).
However, designing such a process for layout is non-trivial, due to the \emph{heterogeneous} nature of the layout sequence, where the tokens representing element types are \emph{categorical} while the tokens representing element coordinates are \emph{ordinal}.
Existing discrete forward processes hardly consider heterogeneous tokens.
Directly applying them to layout data often leads to harsh corruptions, where a layout is changed dramatically at each step (see \cref{noise_absorb,noise_uniform}).
For example, the uniform process in~\cref{noise_uniform} will transition an element type token to a coordinate token, drastically violating the layout semantics.

To realize a mild corruption process for layout, we make three important observations.
\emph{(i) Legality}. 
The transition between type tokens and coordinate tokens will lead to an illegal layout sequence, resulting in an unpredictable change between forward steps.
Hence, it is vital to impose legality during the corruption process.
\emph{(ii) Coordinate Proximity}.
Coordinate tokens are ordinal, and thus transitioning a coordinate token to its proximal tokens (\eg, from $0$ to $1$) will introduce a milder change to a layout compared with transitioning to distant ones (\eg, from $0$ to $127$).
\emph{(iii) Type Disruption}.
Unlike coordinate tokens, type tokens are categorical and do not have particular proximity.
Simply transitioning one type to another may cause abrupt semantic changes to a layout (\eg, from a button to a background image).

Motivated by the above observations, we propose a block-wise transition matrix coupled with a piece-wise linear noise schedule in LayoutDiffusion.
The transition matrix is designed as follows.
First, to achieve legality, we only allow the internal transition between coordinate tokens and that between type tokens.
Second, regarding coordinate proximity, we leverage discretized Gaussian~\cite{d3pm}, where the transition between more proximal tokens takes a higher probability, for the transition between coordinate tokens.
Third, as for type disruption, we introduce absorbing state~\cite{d3pm}.
Each type token either stays the same or transitions to the absorbing state.
To further alleviate type disruption, we propose a piece-wise linear noise schedule to make the transition for element types only occur in the late stage of the forward process.
With above techniques, LayoutDiffusion achieves the mild forward process shown in~\cref{noise_ours}.

Our design also enables LayoutDiffusion to perform certain conditional layout generation tasks in a \emph{plug-and-play} manner without re-training, which has never been explored by previous work.
Specifically, owning to the mild forward process achieved by LayoutDiffusion, its reverse process is to iteratively improve a layout, which naturally supports the task of layout refinement~\cite{rahman2021ruite}.
Besides, as the transition of element types only occurs in the late forward process, LayoutDiffusion will determine the element types in a layout quickly in the reverse process.
Thus, it can perform generation conditioned on types by simply keeping the types fixed and running the reverse process.

In summary, this work makes four key contributions:
\begin{compactenum}

\item We formulate layout generation as a discrete diffusion process, which addresses biased context modeling by iterative refinement from a non-autoregressive model.

\item We design a new diffusion process based on the heterogeneous nature of layout sequence (legality, coordinate proximity and type disruption). It not only better suits layout data but also showcases a promising way of applying diffusion models to other heterogeneous data.

\item We enable certain conditional layout generation tasks in a plug-and-play manner without re-training. 

\item We make extensive experiments and user studies.
LayoutDiffusion outperforms existing methods on all the tasks in terms of most evaluation metrics, even if it is not re-trained for conditional generation tasks. 

\end{compactenum}
\section{Related Work}
\label{sec:related}

\noindent \textbf{Graphic Layout Generation.}
Early work on graphic layout generation has explored classical optimization approaches~\cite{o2014learning,o2015designscape}, as well as generative models such as Generative Adversarial Networks (GANs)~\cite{layoutgan,layoutganpp} and Variational Autoencoders (VAEs)~\cite{layoutvae,canvasvae,vtn,c2f}. 

Recently, inspired by the success of NLP, masking strategies~\cite{blt}, language models~\cite{layouttransformer}, and encoder-decoder architectures~\cite{unilayout} have been studied. 
These approaches represent the layout as a sequence of elements and use Transformer~\cite{vaswani2017attention} as the basic model architecture.
As the placement of one element can depend on any part of a layout, one critical issue in layout generation is \emph{global context modeling}. 
Some previous studies introduce unnatural biases and fail to model global context effectively~\cite{layouttransformer,vtn,unilayout,c2f,canvasvae}. 
They generate the layout sequence in an autoregressive manner, where there is a predefined generation order and the placement of one element can only depend on the generated part of the layout. 
On the other hand, a few other studies consider global context but do not achieve significantly better performance~\cite{layoutgan,layoutganpp,blt}. 
They generate the layout sequence in a non-autoregressive manner, where there is no predefined generation order, and all the tokens are generated in parallel. However, generating a sequence in a single pass is too challenging~\cite{maskpredict}. 
BLT~\cite{blt} explored an iterative refinement mechanism to alleviate the difficulty. 
However, it relies on heuristic rules instead of being learned from the data.
The above limitation motivates us to seek a better model for layout generation. 
We think diffusion models are well-suited.
By multiple rounds of denoising, it naturally takes the layout in the last step as the global context and generates a layout iteratively instead of by a single pass.

Another branch of studies has explored incorporating diverse user constraints into the layout generation~\cite{ndn,rahman2021ruite,zheng2019content,zhou2022composition,li2020attribute}. 
They treat layout generation tasks with different constraints separately, which introduces repetitive training and hinders knowledge sharing across different tasks. 
By utilizing the flexible forward process of diffusion models~\cite{cold_diffusion}, we enable some conditional generation tasks without re-training for the first time, which can be potentially extended to handle more conditional generation tasks.

\noindent \textbf{Diffusion Models for Discreate Data.}
Diffusion models on continuous data have achieved outstanding results~\cite{survey}.
Recently, diffusion models on discrete data are also emerging.
They can be grouped into two categories.
The first category~\cite{diffusionlm,analog,diffuseq} maps discrete data to continuous state space via a learnable or fixed embedding, and then utilizes techniques from classical continuous diffusion models. 
These approaches enable simple technology migration from continuous diffusion models, but make the fine-grained control of the forward corruption process much difficult.
Another category~\cite{argmax,d3pm,vqdiffusion,vqpose,improvedVqpose,tauldr} chooses to directly perform diffusion in discrete state space by modeling the forward corruption process as a random walk between different states. 
This category makes it easy to incorporate domain-dependent structure to the transition matrices and thus enables flexible control of the forward process.
Different to the discrete data (\eg, images and texts) explored by previous work, the layout data studied in this work is heterogeneous by nature.
Thus, we fully consider such characteristic and propose a new transition matrix coupled with noise schedules to achieve a mild corruption process.

\section{Problem Formulation}
\label{formulation}

\noindent\textbf{Graphic Layout.}
A graphic layout $x$ is composed of a set of graphic elements $\left\{\boldsymbol{e}_i\right\}^N_{i=1}$, where $N$ denotes the number of elements.
Each element $\boldsymbol{e}_i$ has an element type $c_i$ and a bounding box indicating its left $l_i$, top $t_i$, right $r_i$, and bottom $b_i$ coordinates.
Following the advanced layout generation methods~\cite{vtn,c2f,layouttransformer,canvasvae,nguyen2021diverse, rahman2021ruite}, we represent an element as a sequence with 5 discrete tokens, i.e., $\boldsymbol{e}_i = \left\{c_i l_i t_i r_i b_i\right\}$, where the continuous bounding box coordinates are uniformly discretized into integers between $[0,K)$. Then, we represent a layout as a concatenation of element sequences:
\vspace{-15px}
\begin{align}\label{eq:layout}
    \mathbf{x} = \left\{\langle \text{sos}\rangle c_1 l_1 t_1 r_1 b_1\|\dots\|c_N l_N t_N r_N b_N\langle \text{eos}\rangle\right\},
\end{align}
where $\langle \text{sos}\rangle$ and $\langle \text{eos}\rangle$ are special tokens indicating the start and end of a sequence, and token $\|$ indicates the separator between any two elements.
Obviously, the layout sequence is \emph{heterogeneous}.
The element type tokens are \emph{categorical}, while the coordinate tokens are \emph{ordinal}.

\noindent \textbf{Graphic Layout Generation.}
In this work, we primarily focus on unconditional layout generation.
Specifically, we learn a generative model $p_\theta(\mathbf{x})$ parameterized by $\theta$, which synthesizes diverse and high-quality graphic layouts.

\section{LayoutDiffusion}
\label{sec:approach}

\begin{figure*}
  \centering
   \includegraphics[width=0.9\linewidth]{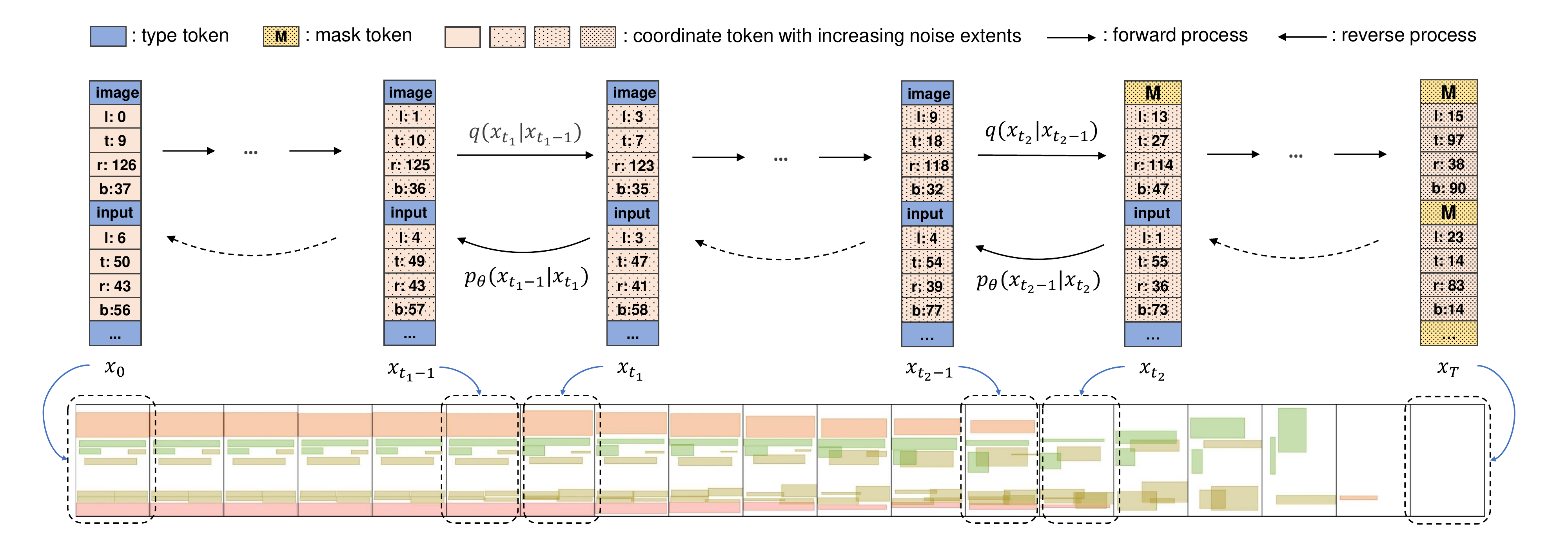}
    \vspace{-10px}
  \caption{An illustration for \textbf{LayoutDiffusion}. In the forward process, the coordinates are mildly corrupted into stationary distribution, and the element types are absorbed into \texttt{MASK} in the late stage. In the reverse process, the element types are first recovered, and then the rough coordinates are gradually refined. For brevity, only two elements are shown, while the other elements and the special tokens are omitted.}
  \label{fig:overview}
  \vspace{-15px}
\end{figure*}

We formulate the layout generation problem as a discrete denoising diffusion process (see \cref{fig:overview}).
It consists of two Markov chains, where the forward process is hand-designed and fixed while the reverse process is parameterized.

Give a real layout $\mathbf{x}_0\sim q(\mathbf{x_0})$, the \emph{forward} process corrupts it into a sequence of increasingly noisy latent variables $\mathbf{x}_{1:T}=\mathbf{x_1},\mathbf{x}_2,\dots,\mathbf{x}_T$,
\begin{align}\label{eq:forward}
    & q(\mathbf{x}_{1:T}|\mathbf{x}_0)  = \prod_{t=1}^T q(\mathbf{x}_t|\mathbf{x}_{t-1}), \\\label{eq:transition}
    & q(x_{t}|x_{t-1}) = x_t \mathbf{Q}_t x_{t-1}.
\end{align}
Here, $x_t$ denotes the one-hot version of a single discrete token in the layout sequence $\mathbf{x}_t$. $\mathbf{Q}_t$ is the transition matrix, where $[\mathbf{Q}_t]_{ij}=q(x_t=j|x_{t-1}=i)$ represents the probabilities that $x_{t-1}$ transitions to $x_t$.
Due to the property of Markov chain, the cumulative probability of $x_t$ at arbitrary timestep from $x_0$ can be derived as $q(x_{t}|x_{0})=\mathbf{x}_t \overline{\mathbf{Q}}_t \mathbf{x}_{0}$, where $\overline{\textbf{Q}}_t=\textbf{Q}_1\textbf{Q}_2\dots\textbf{Q}_t$ (refer to \cite{d3pm} for details).

To generate a layout, the \emph{reverse} process starts with a random noise $\mathbf{x}_T$ and gradually recovers it relying on the learned posterior distribution $p_\theta(\mathbf{x}_{t-1}|\mathbf{x}_t)$,
\begin{align}\label{eq:reverse}
    p_\theta(\mathbf{x}_{0:T}) = p(\mathbf{x}_T)\prod_{t=1}^T p_\theta (\mathbf{x}_{t-1}|\mathbf{x}_t).
\end{align}

In the following, we will introduce how to construct a mild forward process $q(\mathbf{x}_t|\mathbf{x}_{t-1})$ for layout generation (\cref{subsec:forward}), and how to learn the generative model $p_\theta(\mathbf{x}_{0})$ in the reverse process (\cref{subsec:reverse}).

\subsection{Forward Process}
\label{subsec:forward}

In LayoutDiffusion, we propose a block-wise transition matrix $\mathbf{Q}_t$ and a piece-wise linear noise schedule to realize a mild forward process, in which layouts in the neighboring steps do not differ too much and become increasingly disordered as the forward step grows (see \cref{noise_ours}).

The design of the transition matrix and noise schedule stems from our three important observations.
\emph{(i) Legality.}
As defined in~\cref{formulation}, layout sequence has a rigorous format.
Any transition between element type tokens and coordinate tokens will lead to an illegal layout sequence, resulting in a disruptive change between forward steps.
Hence, it is vital to impose sequence legality in the transition matrix.
\emph{(ii) Coordinate Proximity.}
Coordinate tokens in layout sequence are ordinal and have a meaningful proximity.
Transitioning a coordinate token to its proximal tokens (e.g., from $0$ to $1$) will introduce a milder change to a layout, compared with transitioning to distant ones (e.g., from $0$ to $127$).
Thus, it is helpful to encode the proximity prior in the transition matrix.
\emph{(iii) Type Disruption.}
Type tokens are categorical and do not present particular proximity.
Each type of element has its unique coordinate distribution.
For example, a background image tends to have a large size, while a button has a small size.
Transitioning a type to another type may produce an abnormal element (e.g., a button has a large size and is placed at the top-left corner), leading to abrupt changes in layout. 
This is also consistent with the observation from diffusion models on other categorical
data, e.g., latent code and text~\cite{vqdiffusion,d3pm}.
Therefore, it is beneficial to alleviate type disruption in the transition matrix and noise schedule.

\noindent \textbf{Transition Matrices.}
There are three kinds of tokens in the layout sequence, including type tokens (i.e., $c_i$), coordinate tokens (i.e., $l_i$, $t_i$, $r_i$ and $b_i$) and special tokens (i.e., $\langle \text{sos}\rangle$, $\langle \text{eos}\rangle$, $\|$, and $\texttt{PAD}$).
Denote the number of different coordinate tokens and type tokens as $K$ and $C$.
Then, the transition matrix is denoted as $\mathbf{Q}_t \in \mathbb{R}^{V\times V}$, where $V = K+C+4$.

To achieve the legality of the layout sequence, we only allow the internal transition within each kind of tokens.
Thus, $\mathbf{Q}_t$ can be reduced to a block-wise diagonal matrix,
\begin{equation}\label{eq:transition_all}
    \mathbf{Q}_t={
\left[ \begin{array}{ccc}
\mathbf{Q}_t^{\text{coord}} &  & \\
 & \mathbf{Q}_t^{\text{type}} &\\
 & & \mathbf{Q}_t^{\text{spec}}
\end{array} 
\right ]},
\end{equation}
where $\mathbf{Q}_t^{\text{coord}}$, $\mathbf{Q}_t^{\text{type}}$ and $\mathbf{Q}_t^{\text{spec}}$ depicts the probabilities of the internal transition within coordinate tokens, type tokens and special tokens, respectively.

For $\mathbf{Q}_t^{\text{coord}}$, to encode the ordinal proximity, we introduce the discretized Gaussian matrix~\cite{d3pm} for coordinate tokens, which assigns a higher probability to the transition between more proximal tokens,
\begin{equation}
    \left[\mathbf{Q}_t^{\text{coord}}\right]_{ij} = \begin{cases}
\dfrac{\exp\left(-\frac{4|i - j|^2 }{(K-1)^2\beta_t}\right)}{\sum_{n=-(K-1)}^{(K-1)}\exp\left(-\frac{4n^2 }{(K-1)^2\beta_t}\right)}, &i\neq j\\[2.2em]
1 - \sum_{l=0, l\neq i}^{(K-1)} [\mathbf{Q}_t^{\text{coord}}]_{il}, &i = j \\
\end{cases}
    \label{eq:discretized_gaussian_mat}
\end{equation}
where the parameters $\beta_t$ influence the variance of the forward process distributions.

For $\mathbf{Q}_t^{\text{type}}$, to alleviate the type disruption, we choose to transit a type token to a special \texttt{MASK} token instead of another meaningful type token.
Therefore, we introduce the absorbing state transition matrix~\cite{d3pm} for type tokens, 
\begin{equation}
\mathbf{Q}_t^{\text{type}}={
\left[ \begin{array}{cccc}
1-\gamma_t & 0 &\cdots &0\\
0 & 1-\gamma_t &\cdots &0\\
\vdots &\vdots &\ddots &\vdots\\
\gamma_t & \gamma_t &\cdots & 1
\end{array} 
\right ]},
\end{equation}
where $\gamma_t$ indicates the probability that a token is absorbed into a \texttt{MASK} token, and $1-\gamma_t$ is the probability that a token stays unchanged.

For $\mathbf{Q}_t^{\text{spec}}$, as special tokens describe the structure of the layout sequence, any transition between them will lead to an invalid layout sequence. 
Therefore, we choose to disable any transition between them,
\begin{equation}
    \mathbf{Q}_t^{\text{spec}} = \mathbf{I},
\end{equation}
where $\mathbf{I}$ is an identity matrix.

\noindent \textbf{Noise Schedules.} 
An early absorbing of type tokens (i.e., transitioning to \texttt{MASK} token) will bring an abrupt change to the layout.
Hence, to further eliminate type disruption, we choose to make the element type begin to change only in the late stage of the forward process.
Specifically, we design $\overline{\gamma}_t = 1-\prod_{i=1}^t (1-\gamma_i)$ for the cumulative probability $q_t(x_t|x_0)$ as a piece-wise linear function,
\begin{equation}\label{eq:gamma_schedule}
    \overline{\gamma}_t = \begin{cases}
    0, &t< \Tilde{T} \\
(t-\Tilde{T})/(T-\Tilde{T}), &t \ge \Tilde{T} \\
\end{cases}
\end{equation}
Here, $\Tilde{T}$ is the timestep where the absorbing is enabled, and $T$ is the terminal timestep.

Besides, although existing work often uses linear schedule for Gaussian transition process, we choose to use $\beta_t=g/(T-t+\epsilon)^h$ for the transition of coordinate tokens $\mathbf{Q}_t^{\text{coord}}$. 
Here $g$ and $h$ are hyper-parameters, and $\epsilon$ denotes a small positive quantity.
It is generalized from a commonly used noise schedule $1/(T-t+1)$~\cite{2015,d3pm}.
We find that with $h>1$, it achieves a slower and more smooth corruption to the layout in the early forward process, which helps the model in the reverse process better learn the posterior distribution.

\subsection{Reverse Process}
\label{subsec:reverse}
To reverse the forward process, we optimize the generative model $p_\theta(\mathbf{x}_0)$ to fit the data distribution $q(\mathbf{x}_0)$ by minimizing the variational lower bound (VLB)~\cite{d3pm},
\begin{align}\label{eq:vlb}
    \mathcal{L}_{\text{VLB}} = -\log p_\theta(\mathbf{x}_0|\mathbf{x}_1) + 
    D_{\text{KL}}(q(\mathbf{x}_T|\mathbf{x}_0)\| p(\mathbf{x}_T)) \\\nonumber
    + \sum_{t=2}^T D_{\text{KL}}(q(\mathbf{x}_{t-1}|\mathbf{x}_t) \| p_\theta(\mathbf{x}_{t-1}|\mathbf{x}_t)).
\end{align}

Following recent work~\cite{d3pm,vqdiffusion}, we predict $p_\theta(\mathbf{x}_{0}|\mathbf{x}_{t})$ instead of $p_\theta(\mathbf{x}_{t-1}|\mathbf{x}_{t})$, and encourage good predictions of $\mathbf{x}_0$ at each step by combining $\mathcal{L}_{\text{VLB}}$ with an auxiliary objective,
\begin{equation}\label{eq:total_loss}
    \mathcal{L} = \mathcal{L}_{\text{VLB}} - \lambda\log p_\theta(\mathbf{x}_0|\mathbf{x}_t).
\end{equation}

Specifically, we leverage Transformer encoder~\cite{vaswani2017attention} to learn $p_\theta(\mathbf{x}_{0}|\mathbf{x}_{t})$.
Denote the embedding of $i$-th token in the layout sequence $\mathbf{x}_t$ as $\texttt{emb}(x_{t,i})$ and its positional embedding as $p_i$.
Denote the embedding of the timestep $t$ as $\texttt{emb}(t)$.
Then, Transformer takes the aggregation of them, i.e., $\{\texttt{emb}(x_{t,i})+p_i+\texttt{emb}(t)\}_{i=1}^M$, as the input and predicts a new layout sequence $\Tilde{\mathbf{x}}_0 = \{\Tilde{x}_{0,i}\}_{i=1}^M$ as the output. 

In practice, we set an $N$ as the maximum number of elements. During inference, we first sample an element count $n$ from the training set's prior distribution. For constructing $\mathbf{x}_T$, we assign $\texttt{MASK}$ tokens for the type and random coordinate tokens for bounding boxes of the first $n$ elements. For the remaining $(N-n)$ elements, $\texttt{PAD}$ tokens are utilized to ensure a consistent length. By performing denoising from timestep $T$ to $0$, we derive the layout $\mathbf{x}_0$.

\subsection{Enabling Conditional Layout Generation in a Plug-and-Play Manner}
\label{subsec:condition_gen}
Although LayoutDiffusion is trained for unconditional layout generation, it can handle some conditional generation tasks without re-training, which has never been explored by previous work. 
Such a plug-and-play feature of conditional generation is enabled by the design of transition matrices and noise schedules.
In the following, we introduce how LayoutDiffusion achieves it.

\noindent \textbf{Refinement} is a user-oriented layout generation task first posed in RUITE~\cite{rahman2021ruite}, and is recently studied by LayoutFormer++~\cite{unilayout}. 
Its goal is to take a user given flawed layout as input and provide a high-quality layout for the user while maintaining the original design style.
With the proposed transition matrices and noise schedules, a layout is gradually corrupted in the forward process.
With such a forward process, the reverse process learned by LayoutDiffusion is to iteratively improve a layout, which naturally enables refinement.
Specifically, in LayoutDiffusion, we achieve refinement by feeding the flawed layout into the model and then running reverse process from a certain timestep. 
Here the timestep is related to how noisy the input layout is.

\noindent\textbf{Generation Conditioned on Types (Gen-Type)} is also a widely studied conditional layout generation task~\cite{ndn,blt,unilayout} to satisfy the needs of user.
It aims to generate layouts with the given element types.
In LayoutDiffusion, there is no transition between coordinates and types (see \cref{eq:transition_all}).
Besides, with the noise schedule in \cref{eq:gamma_schedule}, the change of the types only occurs in the late forward process.
With the above two mechanisms, LayoutDiffusion will determine the element types in the early reverse steps very quickly and then continue to improve the coordinates in the remaining reverse steps without changing the types (see the transformation of the layout from right to left in \cref{fig:overview}). 
In other words, the generation for coordinates and that for element types are approximately decoupled.
Thus, in LayoutDiffusion, we achieve Gen-Type by feeding in the element types in the early stage and running the reverse process.

\begin{table*}
    \centering
    \setlength{\tabcolsep}{3.5mm}{
    \renewcommand{\arraystretch}{0.95}
    \begin{small}
        \resizebox{\textwidth}{!}{
            \begin{tabular}{llcccccccc}
                \specialrule{1.1pt}{0pt}{1pt}
                                            &                       & \multicolumn{4}{c}{RICO}   & \multicolumn{4}{c}{PubLayNet}                                                                                                                                                                                 \\ \cmidrule(l){3-6} \cmidrule(l){7-10}
                Subtasks                    & \makecell[c]{Methods} & mIoU ($\uparrow$)           &   Overlap ($\rightarrow$)              & Align. ($\rightarrow$)        & FID ($\downarrow$)& mIoU ($\uparrow$)            &   Overlap ($\rightarrow$)            & Align. ($\rightarrow$)        & FID ($\downarrow$)      \\ \midrule
                \multirow{8}{*}{Un-Gen}      
                                            
                                            & LayoutTransformer     &0.587 & \underline{0.542} & {0.037} & 24.320   & 0.359 & \underline{0.0045} & 0.067 & 30.048      \\
                                            & VTN                   &0.336 & 0.561 & 0.477 & 88.115     & 0.312 & 0.221  & 0.207 & 105.909     \\ 
                                            &Coarse2Fine           & 0.360 & 0.676 & \underline{0.128} & 46.483& 0.361 & 0.142  & 0.221 & 50.854     \\ 
                                            &LayoutFormer++           & \underline{0.634} & 0.546 & 0.051 & 20.198  & 0.401 & {0.0010} & {\textbf{0.028}} & 47.082  \\ \cmidrule{2-10}
                                            
                                            &Diffusion-LM$^\diamond$    & {\textbf{0.662}} & 0.631 & 0.184 & 11.448  & {\textbf{0.439}} & 0.0125 & 0.076 & 11.895 \\
                                            &D3PM (absorbing)$^\diamond$ &0.585 & 0.619 & 0.157 & \underline{4.985}  & 0.401 & 0.0427 & 0.075 & 12.218  \\
                                            &D3PM (uniform)$^\diamond$   &0.595 & 0.658 & 0.229 & 5.576   & 0.405 & 0.0571 & 0.099 & \underline{11.212}  \\
                                            &LayoutDiffusion$^\diamond$ (ours)
                                            &0.620 & {\textbf{0.502}} & {\textbf{0.069}} & {\textbf{2.490}}   & \underline{0.417} & {\textbf{0.0030}} & \underline{0.065} & {\textbf{8.625}}  \\
                                             \midrule

                \multirow{8}{*}{Gen-Type}   
                                            & NDN-none   & \underline{0.35}  & 0.55  & 0.56  & 13.76     & 0.31  & 0.17  & 0.35  & 35.67     \\ 
                                            & LayoutGan++                  & 0.298 & 0.620 & 0.261 & 5.954     & 0.297 & 0.148 & 0.124 & 14.875   \\ 
                                            & BLT           & 0.216 & 0.983 & 0.150 & 25.633    & 0.140 & 0.196 & 0.036 & 38.684    \\ 
                                            & LayoutFormer++           & {\textbf{0.377}} & \underline{0.537} & \underline{0.124} & \underline{2.483}   & 0.333 & \underline{0.009} & {\textbf{0.025}} & 10.151    \\ \cmidrule{2-10}
                                            & Diffusion-LM$^\diamond$    & 0.324 & 0.574 & 0.199 & 6.530   & 0.316 & 0.026 & 0.046 & \underline{7.396}        \\
                                            & D3PM (absorbing)$^\diamond$ & 0.337 & 0.594 & 0.192 & 3.506  & 0.333 & 0.058 & 0.057 & 8.858 \\
                                            & D3PM (uniform)$^\diamond$   & 0.317 & 0.621 & 0.218 & 5.771  & {\textbf{0.351}} & 0.063 & 0.064 & 8.275      \\
                                            
                                            & LayoutDiffusion$^\diamond$ (ours) & {0.345} & {\textbf{0.491}} & {\textbf{0.124}} & {\textbf{1.557}}   & \underline{0.343} & {\textbf{0.005}} & \underline{0.029} & {\textbf{3.731}} \\
                                             \midrule

                \multirow{5}{*}{Refinement}  
                                            & RUITE      &\underline{0.658} & \underline{0.492} & 0.177 & 7.926 & 0.637 & 0.0375 & 0.073 & 7.890  \\ %
                                            & LayoutFormer++           & 0.656 & 0.503 & \underline{0.141} & \underline{3.666} & \underline{0.642} & \underline{0.0126} & \underline{0.042} & \underline{2.937}  \\ \cmidrule{2-10}
                                            & Diffusion-LM$^\diamond$    & 0.621 & 0.499 & 0.181 & 8.578 & 0.573 & 0.0408 & 0.140 & 13.985 \\
                                            & D3PM (uniform)$^\diamond$   & 0.568 & 0.526 & 0.266 & 8.407 & 0.564 & 0.0515 & 0.110 & 14.553 \\
                                            & LayoutDiffusion$^\diamond$ (ours) &{\textbf{0.719}} & {\textbf{0.469}} & {\textbf{0.102}} & {\textbf{0.549}} & {\textbf{0.660}} & {\textbf{0.0079}} & {\textbf{0.035}} & {\textbf{2.045}} \\
                                            \midrule

              Real Data  &&-& 0.466 & 0.093 & -  & - & 0.0031 & 0.022 & -     \\
                \specialrule{1.1pt}{1pt}{0pt}
            \end{tabular}}
    \end{small}}
     \vspace{-5px}
        \caption{Quantitative results. Methods with $\diamond$ are diffusion-based, which achieve conditional generation (i.e., Gen-Type and Refinement) in a \textit{plug-and-play} manner, while other methods require re-training for each subtask. The best and the second best values of each metric are \textbf{bold} and \underline{underlined} respectively. For mIoU, the higher the score, the better the performance (indicated by $\uparrow$). For Overlap and Align, the closer to real data, the better (indicated by $\rightarrow$). For FID, the lower the score, the better the performance (indicated by $\downarrow$).
        }
    \label{Tab:quantitative_results}
 \vspace{-15px}
\end{table*}

\section{Experiments}
\label{sec:experiments}
\subsection{Setups}

\noindent \textbf{Datasets.} 
We employ two widely-used public datasets of graphic layouts. 
\textit{RICO}~\cite{rico} is a dataset of user interface designs for mobile applications, which contains 66K+ UI layouts with 25 element types. 
\textit{PublayNet}~\cite{publaynet} includes 360K+ annotated scientific document layouts with 5 element types. 
Both datasets contain a few over-length entries.
We filter out the layouts longer than 20 elements as in LayoutFormer++~\cite{unilayout}. 
Then, we split the filtered data into a training, validation, and test set by 90\%, 5\%, and 5\%.

\noindent \textbf{Baselines.} 
First, we compare LayoutDiffusion with leading approaches for layout generation.
Specifically, we compare against \textit{LayoutTransformer}~\cite{layouttransformer}, \textit{VTN}~\cite{vtn}, \textit{Coarse2Fine}~\cite{c2f}, and \textit{LayoutFormer++}~\cite{unilayout} on unconditional generation (UGen); against \textit{NDN-none}~\cite{ndn}, \textit{LayoutGan++}~\cite{layoutganpp}, \textit{BLT}~\cite{blt}, and \textit{LayoutFormer++}~\cite{unilayout} on generation conditioned on type (Gen-Type); against \textit{RUITE}~\cite{rahman2021ruite}, and \textit{LayoutFormer++}~\cite{unilayout} on refinement. 
Moreover, we compare LayoutDiffusion with the existing diffusion models that do not consider the characteristics of layouts.
\textit{Diffusion-LM}~\cite{diffusionlm} maps discrete data to continuous state space, while \textit{D3PM (uniform)}~\cite{d3pm} and \textit{D3PM (absorbing)}~\cite{d3pm} perform diffusion in discrete state space using different transition matrices.

\noindent \textbf{Implementation Details.} 
We set the weight of auxiliary loss as $\lambda=0.0001$ (see \cref{eq:total_loss}).
In the training, we set the timestep as  $T=200$; in the inference, we set the timesteps $T_{\text{UGen}}=200$, $T_{\text{Gen-Type}}=160$, and $T_{\text{Refine}}=50$ for different generation tasks.
For the layout sequence (see \cref{eq:layout}), we arrange elements in the alphabetical order of the type, and each token is embedded with $d=128$ dimensions.
For the denoising network $p_\theta(\mathbf{x}_0|\mathbf{x}_t)$, we apply a 12-layer Transformer encoder with 12 attention heads.
We train the model using AdamW optimizer~\cite{adam} with $2\sim4$ NVIDIA V100 GPUs. 
We also employ importance timestep sampling~\cite{improvedDdpm} during the training. 
See Supplementals for more details.

\noindent \textbf{Evaluation Metrics.}
We adopt four metrics to measure the performance comprehensively. Among them, Frechet Inception Distance (\textbf{FID}) measures the overall performance, while Maximum Interaction over Union (\textbf{mIoU}), Alignment (\textbf{Align.}) and \textbf{Overlap} measure the quality from a specific aspect.
Specifically, \emph{FID} computes the distance between the distribution of the generated layouts and that of real layouts. Following the previous practice~\cite{ndn,layoutganpp}, we train a classification-based neural network to get the feature embedding for the layout.
\emph{mIoU} calculates the maximum IoU between bounding boxes of the generated layouts and those of the real layouts with the same type set~\cite{layoutganpp}.
\emph{Align.} measures whether the elements in a generated layout are well-aligned, either by center or by edges. 
In addition to the original implementation~\cite{ndn}, a normalization over the number of elements is applied.
\emph{Overlap} measures the overlapping area between elements in the generated layout. 
Following LayoutFormer++~\cite{unilayout}, we ignore normal overlaps, \eg, elements on top of the background.

\begin{figure*}
  \centering
   \includegraphics[width=1\linewidth]{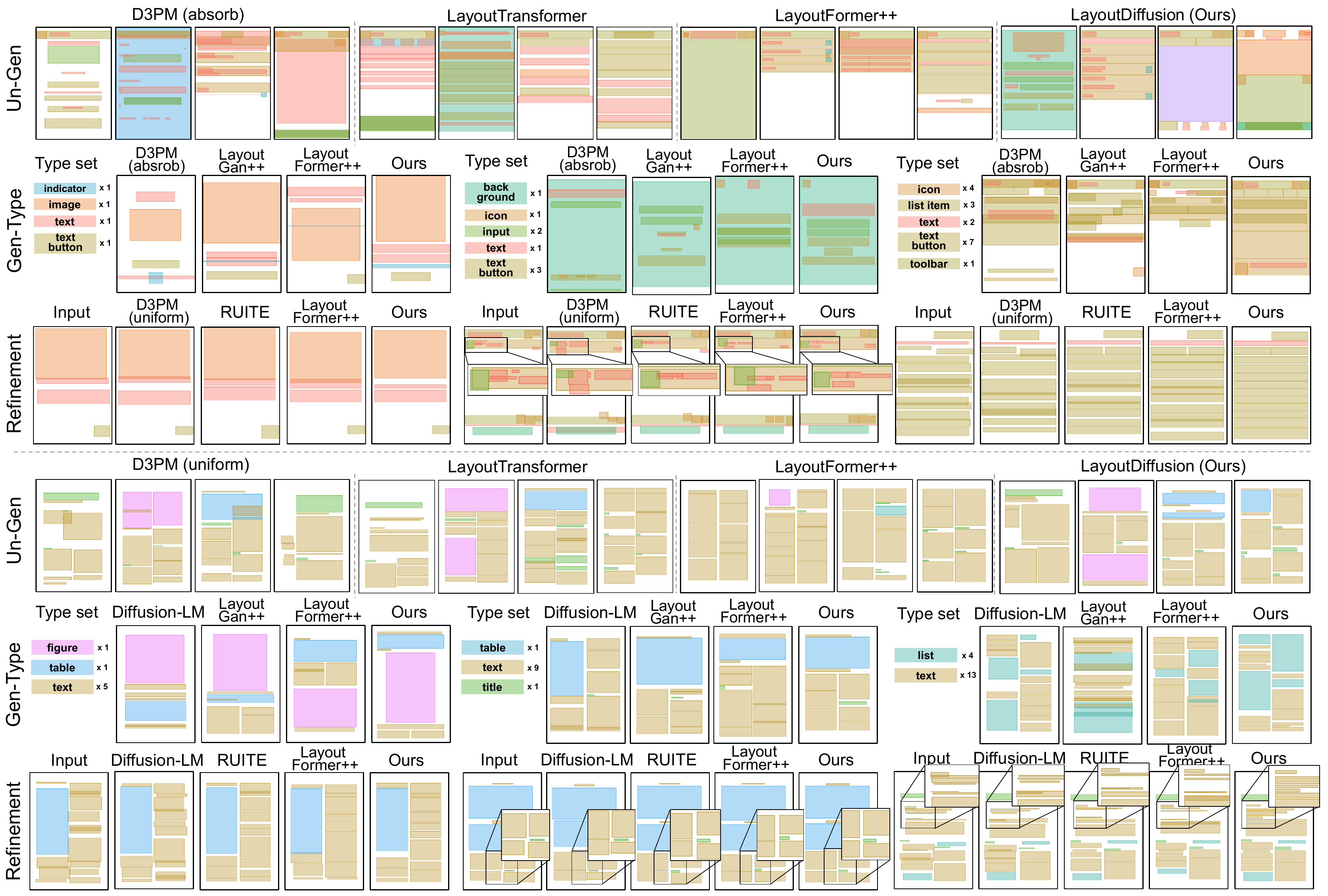} %
    \vspace{-20px}
  \caption{Qualitative comparison against strongest baselines selected by FID (better view in color and 2$\times$ zoom). 
  The first three row is for RICO and the last three is for PubLayNet.
  LayoutDiffusion generates high-quality and diverse layouts. Layouts from LayoutFormer++ either lack diversity (Un-Gen) or are flawed (Gen-Type and Refinement). Layouts from other methods misalign and overlap frequently.}
  \label{qualitative_ungen}
  \vspace{-15px}
\end{figure*}

\subsection{Comparison with Existing Approaches for Layout Generation}

\noindent \textbf{Quantitative Analysis.}
In \cref{Tab:quantitative_results}, the methods without the symbol $\diamond$ are existing approaches for layout generation.

First, we compare FID as it is an overall metric for generation performance. 
LayoutDiffusion achieves significantly better FID scores than all other methods.
For example, on Un-Gen, LayoutDiffusion achieves \textbf{2.490} and \textbf{8.625} on RICO and PubLayNet datasets, respectively, while the best existing work only achieves 20.198 and 30.048.

Furthermore, we examine individual metrics, including mIoU, Overlap, and Align., each of which measures the quality from a specific aspect. 
LayoutDiffusion is in the top two on almost every metric and frequently achieves the best performance.
On the contrary, existing approaches may perform well on a certain metric but fail on the other individual metrics and the overall metric (i.e., FID), indicating that LayoutDiffusion is a well-rounded approach.
For example, on Un-Gen, LayoutTransformer has a good overlap score, but it does not perform well on Align., mIoU and FID; LayoutFormer++ has the best Align. for PubLayNet, but underperforms in mIoU, overlap and FID.

Moreover, on conditional generation tasks (i.e., Gen-Type and Refinement), the above observations still hold, even though LayoutDiffusion is \textit{not re-trained} on these tasks (see \cref{subsec:condition_gen}) while existing approaches are re-trained.

\noindent \textbf{Qualitative Analysis.}
\cref{qualitative_ungen} shows qualitative results. 
On Un-Gen, LayoutDiffusion generates diverse and high-quality layouts. 
In contrast, LayoutTransformer mainly suffers from incorrect spacing and overlap, and LayoutFormer++ is deficient in diversity.
For example, for LayoutFormer++, most layouts on RICO contain a top toolbar and several list items, and most layouts on PublayNet are double-columned and have many texts.
Besides, on Gen-Type and Refinement, LayoutDiffusion outperforms other methods (\eg., alignment, overlap and spacing) while it is not re-trained.
For more qualitative results, please refer to \cref{sec:suppl_quali}.

\noindent \textbf{User Study.}
On each task, we select the best two baselines by FID for the user study.
We design two kinds of evaluation.
One is \emph{quality} evaluation. We show three layouts from three models respectively (two from baselines and one from LayoutDiffusion) and invite the user to choose which one has the best quality (\eg, more plausible overall structure and pleasing details). 
Another one is \emph{diversity} evaluation. We show three sets of layouts from three models respectively, where each set contains five layouts from the same model.
Then, we invite the user to choose which set has the most diverse layouts.
For Refinement, we do not conduct diversity evaluation as it is not necessary for this scenario.

\begin{figure*}
  \centering
   \includegraphics[width=0.925\linewidth]{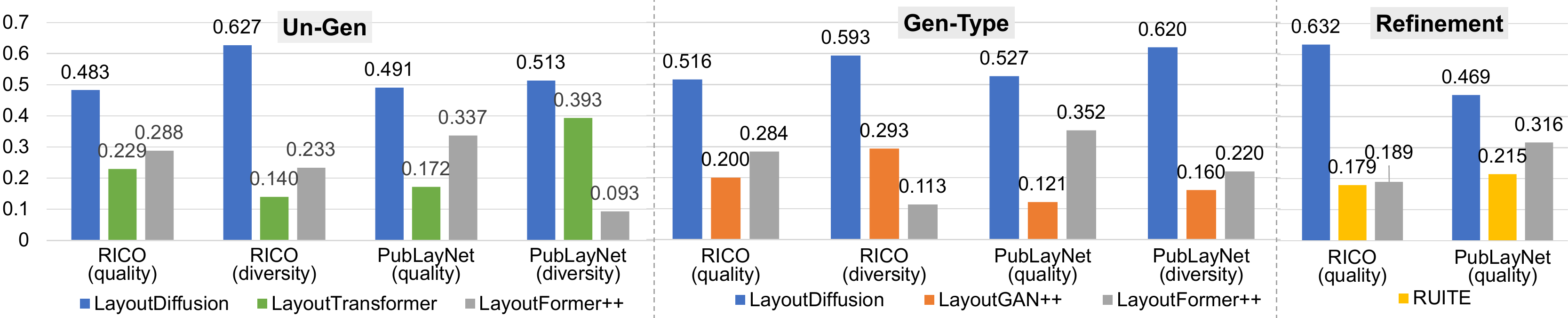}
    \vspace{-10px}
  \caption{Results of the user study. For each model, we count how many people prefer the layouts generated from this model. The study shows that the results generated by LayoutDiffusion were favored by users over the other methods, particularly in terms of diversity.}
  \label{user_study}
  \vspace{-15px}
\end{figure*}

\cref{user_study} shows the results. 
Across different datasets, tasks and evaluation modes, there are 10 groups of user studies in total, in each of which we invite 15 people and everyone labels 50 groups of layouts.
The user study shows that LayoutDiffusion outperforms other methods significantly.

\begin{figure}
  \centering
   \includegraphics[width=0.9\linewidth]{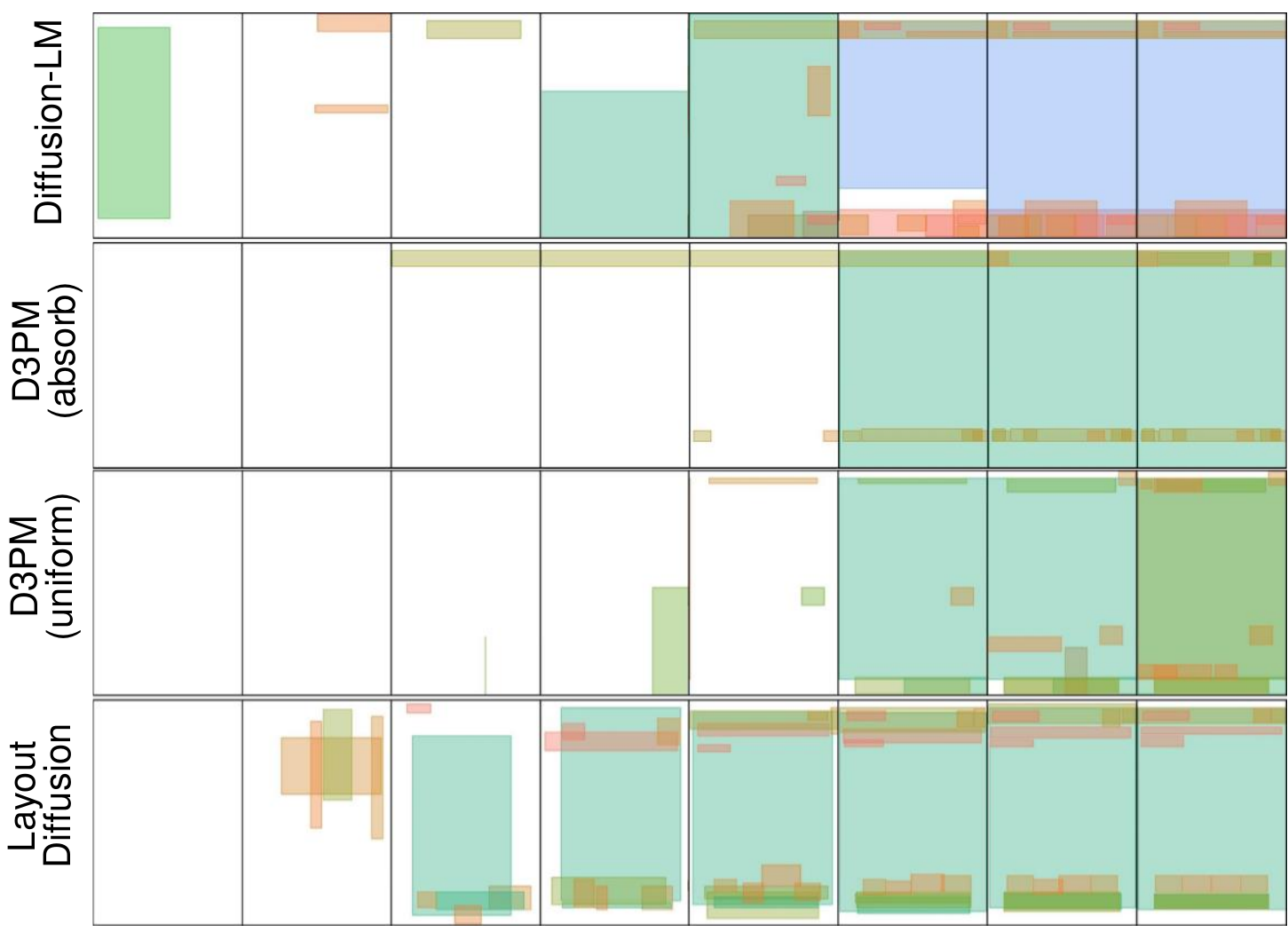}
    \vspace{-10px}
  \caption{Reverse denoising process for unconditional generation on RICO (from left to right). Each row is for one model. The blank page is used when the generated layout sequence is invalid.} 
  \label{fig:case_study}%
  \vspace{-10px}
\end{figure}

\subsection{Comparison with Traditional Diffusion Models}
\noindent \textbf{Quantitative Analysis.}
In \cref{Tab:quantitative_results}, methods marked with $\diamond$ are traditional diffusion models. 
They are originally proposed for other generation tasks (\eg, image and text).
We adapt them for layout generation.
On Un-Gen, LayoutDiffusion achieves the best performance on most metrics.
On Gen-Type and Refinement, while traditional diffusion models can be used for conditional generation tasks without re-training, their performance is usually worse than existing methods for layout generation (\eg, LayoutFormer++), not to mention LayoutDiffusion.
These observations demonstrate that our consideration of the heterogeneous nature of layout data is critical for both achieving good performance and realizing plug-and-play conditional generation.

\noindent \textbf{Qualitative Analysis.}
\cref{qualitative_ungen} shows a qualitative comparison with the best traditional diffusion model (selected by FID).
LayoutDiffusion consistently generates better layouts, \eg, better alignment and less overlap.
Moreover, \cref{fig:case_study} compares the reverse denoising processes of different diffusion models.
LayoutDiffusion quickly generates a draft layout and then gradually refines it to a pleasing layout, while other diffusion models take many steps to generate a rough layout and fewer steps for iterative refinement, which may limit the modeling of precise relationships between elements, such as strict alignment and no overlap.

\subsection{Ablation Studies and Discussions} %
\label{sec:ablation}

\begin{table}[t]
  \centering

\renewcommand{\arraystretch}{0.75}
\begin{small}
  \begin{tabular}{@{}lccccc@{}}
    \toprule
    Method & mIoU$\uparrow$           &   Overlap$\rightarrow$          & Align.$\rightarrow$       & FID$\downarrow$ \\
    \midrule
    Uniform $\mathbf{Q}_t^{\text{coord}}$ &0.614 & 0.633 & 0.119 & 3.174  \\
    Absorbing $\mathbf{Q}_t^{\text{coord}}$ &0.636 & 0.607 & 0.114 & 3.672  \\
    Uniform $\mathbf{Q}_t^{\text{type}}$ &0.593 & 0.478 & 0.139 & 2.534 \\ 
    \midrule
    Linear $\overline{\gamma}_t$ & 0.580 & 0.522 & 0.156 & 2.846  \\
    Linear $\beta_t$ & 0.589 & 0.517 & 0.202 & 2.598  \\ \midrule
    LayoutDiffusion & 0.620 & 0.502 & 0.069 & 2.490   \\ \midrule
    Real Data & - &0.466&0.093&-\\
    \bottomrule
  \end{tabular}
  \end{small}
  \vspace{-5px}
   \caption{Ablation studies on RICO with unconditional generation.}
  \label{ablation}
  \vspace{-7.5px}
\end{table}

\noindent \textbf{Transition Matrices.} 
Our transition matrices are designed by considering three critical factors, i.e., legality, coordinate proximity and type disruption (see \cref{subsec:forward}).
We remove the technique corresponding to each factor.
First, without considering the legality, the techniques for the other two factors cannot be applied.
Thus, LayoutDiffusion degrades to D3PM (absorbing or uniform).
\cref{Tab:quantitative_results} shows that LayoutDiffusion achieves better performance.
Second, to ignore coordinate proximity, we use uniform or absorbing transition for coordinate tokens, denoted as Uniform $\mathbf{Q}_t^{\text{coord}}$ and Absorbing $\mathbf{Q}_t^{\text{coord}}$ in \cref{ablation}.
LayoutDiffusion outperforms these two variations on most metrics, especially Overlap and FID.
Third, to study type disruption, we use uniform transition for type tokens, denoted as Uniform $\mathbf{Q}_t^{\text{type}}$ in \cref{ablation}.
LayoutDiffusion outperforms this variation, where the improvement of mIoU is most significant.

\noindent \textbf{Noise Schedules.} 
To make the corruption of element types occur throughout the forward process, we set $\Tilde{T}$ as $0$ for $\overline{\gamma}_t$ (see \cref{eq:gamma_schedule}), which results in a linear schedule (denoted as Linear $\overline{\gamma}_t$ in \cref{ablation}).
We replace the noise schedule for $\beta_t$ with the original linear schedule in previous work~\cite{d3pm} (denoted as Linear $\beta_t$ in \cref{ablation}).
LayoutDiffusion outperforms both of them on most metrics, especially mIoU and Align.

\begin{table}[t]
\centering
\setlength{\tabcolsep}{1mm}{
\begin{small}
\begin{tabular}{|l|c|c|c|c|c|}
\hline
 \multicolumn{1}{|c|}{\multirow{2}{*}{FID$\downarrow$}}    & \multicolumn{5}{c|}{Number of Steps}\\ \cline{2-6}
      & 100    & 200    & 500    & 1000   & 2000   \\ \hline
Diffusion-LM     & 17.759 & 17.164 & 11.984 & 11.741 & 11.448 \\ \hline
D3PM (absorbing) & 7.464  & 6.102  & 6.091  & 4.985  & 5.110      \\ \hline
D3PM (uniform)   & 6.986  & 6.910  & 5.351   & 5.575  & 5.239     \\ \hline
LayoutDiffusion  & 3.875  & 2.490  & 2.387      & 2.295      & 2.360      \\ \hline
\end{tabular}
\end{small}
}
 \vspace{-5px}
\caption{Ablation study on timesteps for diffusion models. All the experiments are on RICO with unconditional generation. The training and inference steps are set as the same. $2000$ and $1000$ are default settings used in Diffusion-LM~\cite{diffusionlm} and D3PMs~\cite{d3pm}.}
\label{tbl:speed}
\vspace{-17.5px}
\end{table}

\noindent \textbf{Timesteps.} 
\cref{tbl:speed} shows FID of diffusion models with different timesteps.
While $200$ steps are enough for LayoutDiffusion, the performance of Diffusion-LM, D3PM (absorbing), and D3PM (uniform) saturates at $500$, $1000$, and $500$ timesteps respectively.
Besides, even the fastest version of LayoutDiffusion surpasses all the other diffusion models.

\noindent \textbf{Additional Analyses.} For a detailed discussion on the diversity of the generated layouts, see \cref{sec:selfsim}. Experiments on additional settings of conditional generation are available in \cref{more cond}. Ablation studies concerning conditional generation tasks, noise schedules, and sequence ordering can be found in \cref{more ablation}.

\section{Conclusion}
\vspace{-2.5px}
In this work, we propose \textit{LayoutDiffusion} to improve graphic layout generation by discrete diffusion models. 
The core of our method lies in realizing a mild forward process by considering the heterogeneous characteristics of the layout. 
Our method also enables two conditional generation tasks without re-training. 
Experiments demonstrate the superiority of LayoutDiffusion over leading approaches for layout generation and existing diffusion models.
In the future, we plan to incorporate diverse conditions~\cite{nichol2021glide,ho2022classifier} in LayoutDiffusion.
Besides, we will also explore how to extend LayoutDiffusion to handle other heterogeneous data.

\section{Acknowledgement}
We would like to thank Zhaoyun Jiang for helpful discussions. 
We also appreciate Zhaoyun Jiang's support in providing the pre-processed datasets and pretrained models for the FID evaluations.

{\small
\bibliographystyle{ieee_fullname}
\bibliography{egbib}
}
\clearpage
\appendix

\onecolumn

\begin{spacing}{1}
\tableofcontents
\end{spacing}

\newpage

\section{Additional Implementation Details}

\subsection{More Implementation Details for LayoutDiffusion}

\begin{figure}[th]
\centering

   \includegraphics[width=0.35\linewidth]{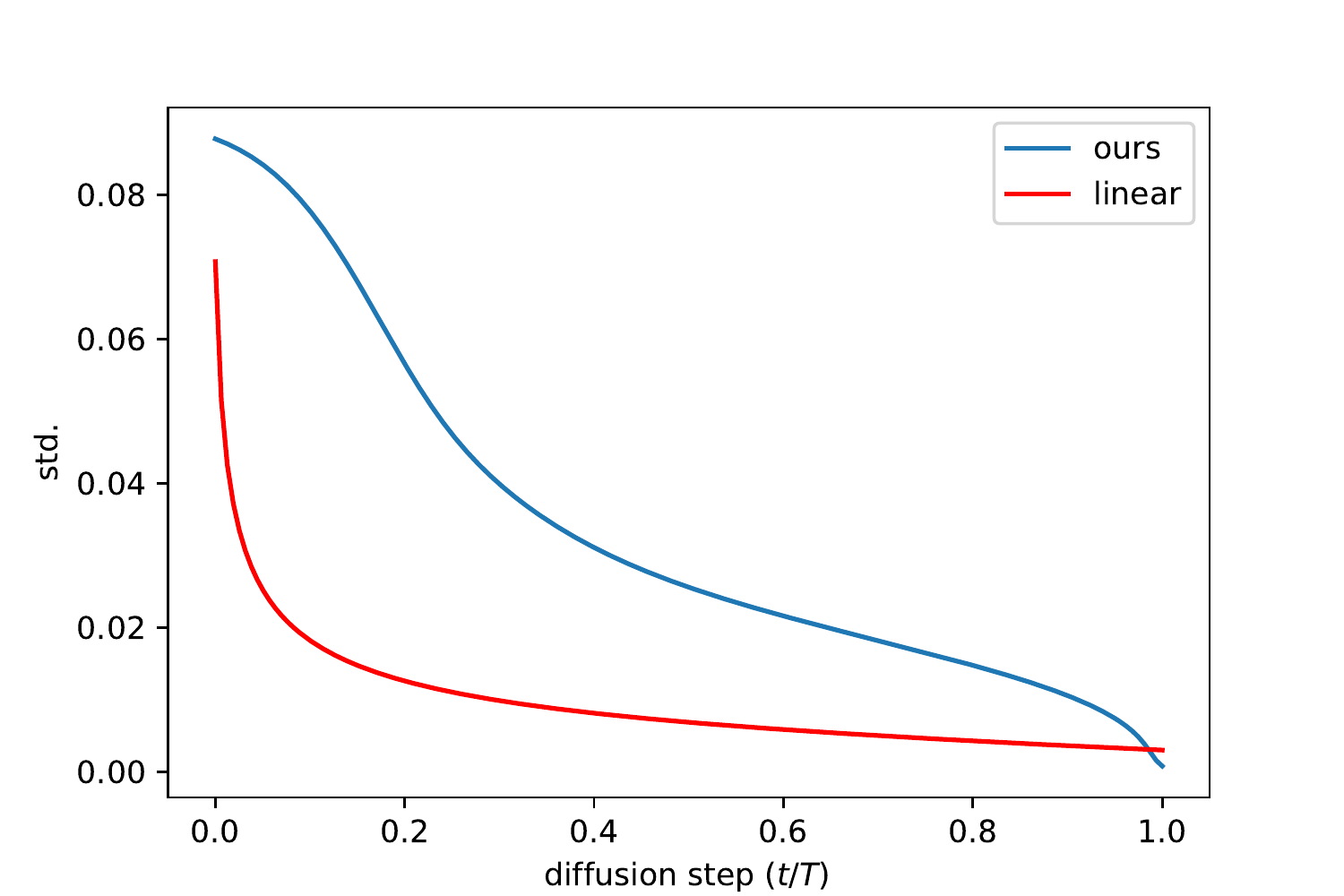}%
\vspace{-5px}
   \caption{Standard derivation of the cumulative discretized Gaussian matrix $\overline{\mathbf{Q}}_t^{coord}$ of our proposed $\beta_t$ and the original linear schedule. }
\vspace{-5px}%
   \label{fig:noise schedule} 
\end{figure}

\noindent \textbf{Noise Schedule.} We investigate the effectiveness of our proposed schedule $\beta_t=g/(T-t+\epsilon)^h$ for the discretized Gaussian transition matrix by comparing with the original linear schedule $\beta_t=bt/T$ used in~\cite{d3pm}. 
Notably, here $\beta_t$ is not a variance term bounded in $[0,1]$\footnote{In fact, as $\beta_t$ tends to positive infinity, $\mathbf{Q}_t^{coord}$ will approach a transtion matrix for uniform noise as described by Sohl-Dickstein \etal~\cite{2015}.}, and with the growth of forward steps, the cumulative matrix $\overline{\mathbf{Q}}_t^{coord}$ converges to uniform distribution.
Thus, we attempt to analyse the noise process by observing the standard derivation of the cumulative matrix.
A higher std. indicates a more sparse matrix and hence a lower transition probability to other coordinate tokens. 
As shown in~\cref{fig:noise schedule}, 
our schedule presents a gentler noising process and a more stable convergence state compared to the original linear schedule (as suggested by a higher std. at the beginning of the forward process, and a lower std. at the very end of the process).

\vspace{-11px}

\begin{figure}[th]
  \centering
   \includegraphics[width=1\linewidth]{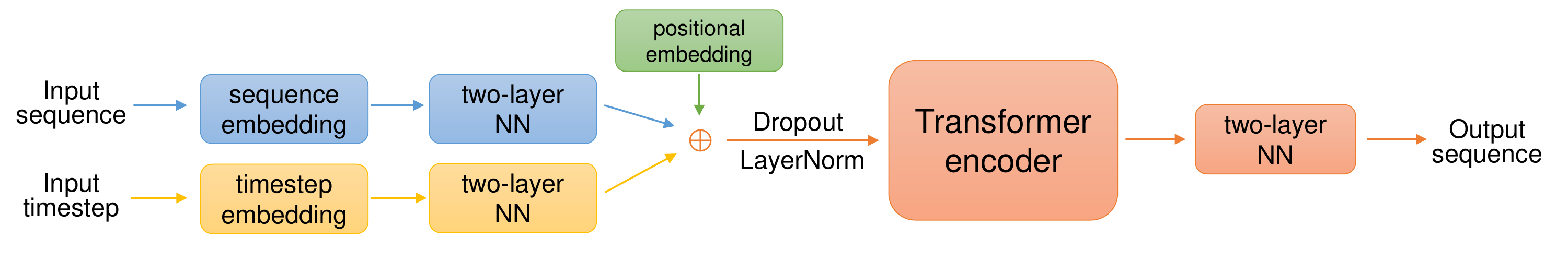}
   \vspace{-10px}
\caption{Model architecture of $p_{\theta}(x_0|x_t)$.}
\vspace{-5px}
   \label{model arch} 
\end{figure}

\noindent \textbf{Denoising Model.}
We present the model architecture as in~\cref{model arch}. We embed the input sequence, input timestep, and positions with 768, 128, and 768 dimensions respectively. The dropout rate is set as 0.1. For the transformer encoder, we simply adopt the same hyperparameters of BERT-base~\cite{bert} encoder, i.e., 12 layers, 12 attention heads, 768 hidden size, and 3,072 dimensions for the feed forward layer.

\noindent \textbf{Hyperparameters.} We train our model using AdamW optimizer~\cite{adam} with $lr=0.00004$, $betas=(0.9,0.999)$, and zero $weight\_decay$. We also apply an exponential moving average (EMA) over model parameters with a rate of $0.9999$.
We set the batch size as 64. For RICO~\cite{rico}, we train the model with 2 V100 GPUs for 175,000 steps to achieve the best results; and for PublayNet~\cite{publaynet}, we train the model with 4 V100 GPUs for 350,000 steps.
For the schedule of type tokens, we set $\Tilde{T}=160$ in~\cref{late absorb}. For the schedule of coordinate tokens, we set $T=\Tilde{T}=160$, $g=12.4$, $h=2.48$, and $\epsilon=0.0001$ in $\beta_t=g/(T-t+\epsilon)^h$. We also provide the hyperparameters for the implementation of different timesteps apart from 200 steps (we implement these variants in the \cref{tbl:speed} of the main paper), as shown in~\cref{tbl:schedule}.

\begin{table}[h]
\centering
\setlength{\tabcolsep}{3mm}{
  \begin{tabular}{lcc}
    \toprule
    Total timesteps &  $\Tilde{T}$ for type schedule &   $\beta_t$ for coordinate schedule \\
    \midrule
    100 &80 & $20.0/(80-t+0.0001)^{2.96}$   \\
    200 &160 & $12.4/(160-t+0.0001)^{2.48}$   \\
    500 &400 & $6.2/(400-t+0.0001)^{2.00}$  \\ 
    1000 & 800 & $3.5/(800-t+0.0001)^{1.76}$   \\
    2000 & 1600 & $2.0/(1600-t+0.0001)^{1.52}$   \\ 
    \bottomrule
  \end{tabular}}
\caption{Hyperparameters for variants of different timesteps.}
\vspace{-5px}
\label{tbl:schedule}
\end{table}

\subsection{Implementation of other Diffusion Baselines}
\noindent \textbf{Diffusion-LM~\cite{diffusionlm}.} We implement Diffusion-LM based on the official repository\footnote{\url{https://github.com/XiangLi1999/Diffusion-LM}}. We realize the layout generation task via Diffusion-LM by feeding the layout as a sequence and reconverting the output sequence to a layout. For the hyperparameters, we simply adopt the default setting that Diffusion-LM adopts on E2E~\cite{e2e} dataset, since only relatively small vocabulary size ($\sim 150$) is required for the tokens that represent layout sequence.

For conditional generation tasks, we apply the similar idea as in LayoutDiffusion. Specifically, for Gen-Type, we fix the type tokens by feeding the target in each timestep and run the whole reverse denoising process. For refinement task, we embed the layout sequence and set the embedded latent as the input of start timestep $T_{\text{refine}}$, and then run the remaining reverse process with type fixed.
For the choice of $T_{\text{refine}}$, we traverse through [250,500,750,1000,1250,1500] of the total 2000 steps, and find the best result is achieved when $T_{\text{refine}}=1000$.

\vspace{5px}
\noindent \textbf{D3PM uniform~\cite{d3pm}.} We implement D3PMs based on both the official repository of D3PMs\footnote{\url{https://github.com/google-research/google-research/tree/master/d3pm}} and the official repository of another method concerning discrete diffusion model, i.e., VQ-Diffusion\footnote{\url{https://github.com/microsoft/VQ-Diffusion}}, since our implementation is based on PyTorch~\cite{paszke2019pytorch} rather than JAX~\cite{jax2018github}. We realize the layout generation task in a similar way as LayoutDiffusion, i.e., feeding the layout as a sequence of tokens and treating each token as a discrete state.

For the hyperparameters of the diffusion framework, we set the total diffusion timesteps $T=1000$, the schedule $\beta_t=(T-t+1)^{-1}$, and the auxiliary loss weight $\lambda=0.0001$, all follow the setting as reported in D3PMs paper. For the denoising model, we apply the similar model as in LayoutDiffusion and Diffusion-LM for a fair comparison.

For conditional generation tasks, we implement the Gen-Type and refinement the same way as in our implementation of Diffusion-LM, since both methods are replace-based diffusion methods. For the start timestep for refinement task $T_{\text{refine}}$, we sweep from [200,400,600,800] of the total 1000 steps, and find $T_{\text{refine}}=400$ is the optimal choice.

\vspace{5px}
\noindent \textbf{D3PM absrobing~\cite{d3pm}.} We implement the D3PM (absorbing) the same way as in D3PM (uniform). All settings except for the model are also referenced from the original D3PMs paper.

One major difference lies in the implementation of conditional generation tasks. 
For the Gen-Type task, we apply the similar idea as in LayoutDiffusion. To be more specific, we feed the given type set at the beginning step $T$, and run the whole reverse process. 
It is noteworthy that, for D3PM (absorbing), all coordinates start to recover strictly from timestep $T$.
Hence, it cannot save steps as the same strategy in LayoutDiffusion by picking a timestep $T_{\text{Gen-Type}}$ which is smaller than $T$. 
Besides, in the reverse process of D3PM (absorbing), the sampled tokens cannot transition into \texttt{MASK} or other tokens, 
thus, it is unable to perform the refinement task as in LayoutDiffusion.

\subsection{Settings on Conditional Generation Tasks}
\label{sec:details cond tasks}
We present below the settings on conditional generation tasks in the main paper.
For further experiments on conditional generation tasks, please refer to~\cref{more cond}.
\vspace{2.5px}

\noindent \textbf{Gen-Type.}
We follow the convention in \cite{blt,unilayout}.
To be more specific, for a layout in the test set, we extract its type set as the input and let the model generate the bounding box attributes of each element .

\vspace{2.5px}
\noindent \textbf{Refinement.} 
In the real scenario, the noise level of the user-given flawed layout cannot be known in advance.
Besides, different flawed layouts may have different noise levels.
To simulate the real scenario, we improve the setting used in RUITE~\cite{rahman2021ruite}.
Specifically, the setting in RUITE is to construct a test set by adding random noises to the position and size of each element, where the noise is sampled from a normal distribution with mean $0$ and the standard variance $0.01$.
In our improved setting, we modify the standard variance of the noise to be uniformly sampled from $[0.005,0.01,0.015,0.02,0.025]$. 
Besides, for the baselines, we train the model with input noise of 0.01 standard variance; for LayoutDiffusion, we apply the same inference steps $T_{\text{refine}}$ for different levels of noise, unlike the settings in~\cref{sec:suppl_refine}.

\subsection{Details about Classification Model for FID Evaluation}
We use the same layout classification model as LayoutGan++~\cite{layoutganpp} and LayoutFormer++~\cite{unilayout}. Specifically, the model includes an encoder and a decoder, both of which are based on the Transformer architecture.
The encoder takes in bounding box coordinates and corresponding labels and produces a feature representation, while the decoder uses the feature representation to predict the class probabilities and bounding box coordinates for each layout element.
We implement the model based on the official repository of LayoutGan++\footnote{\url{https://github.com/ktrk115/const_layout}} and train it using the methods described in their paper.
Our evaluation results using this model are consistent with those reported in the LayoutFormer++.

\section{Discussion on Diversity} %
\label{sec:selfsim}
\subsection{Metric SelfSim}
Diversity is a key but often overlooked aspect in layout generation tasks. In this section, we propose a new metric called \emph{SelfSim} to measure the self-similarity of generated layouts, which serves as an indicator of diversity. The intuition behind this metric is that more diverse generated layouts should be less self-similar. Specifically, we calculate the average Intersection over Union (IoU) between any pairs of generated layouts with the same set of element types.

\subsection{Algorithm of SelfSim}
Inspired by the metrics for evaluating diversity in NLP (\eg, diverse 4-gram~\cite{div4} and self-BLEU~\cite{selfB}), we propose to assess the diversity of generated layouts by measuring the self-similarity of the generated layout set. Specifically, we partition the generated layouts into different subsets based on their type sets and then count the similarity of the layouts within each subset. The similarity is calculated by averaging the intersection of union (IoU) of the bounding boxes for each pair of layouts in the subset.
We present the algorithm for calculating SelfSim as in \cref{algo_selfsim}.

\IncMargin{1em}
\begin{algorithm}[h]
\label{algo_selfsim}
	\caption{Calculation of the Self-Similarity score}
    \vspace{1px}
	\KwIn{A set of graphic layouts $\mathbb{X}=\{\mathbf{x}_1,\mathbf{x}_2,\dots,\mathbf{x}_m\}$} 
	\vspace{1px}
	\KwOut{The Self-Similarity score of the given layout set}
	 \BlankLine 
	 
	 \emph{partition $\mathbb{X}$ by the layouts' \textbf{type set}, denote the partition as $\mathbb{P}=\{\mathbb{X}_1,\mathbb{X}_2,\dots,\mathbb{X}_n\}$}\; 
	 \For(\tcc*[f]{traverse each subset $\mathbb{X}_i \in \mathbb{P}$}){$i\leftarrow 1$ \KwTo $n$}{ 
	 	\emph{count the number of elements in subset $\mathbb{X}_i$, denoted as $l_i$}\; 
	 	\If(\tcc*[f]{only one layout in the subset}){$l_i=1$}{set the Self-Similarity score of subset $\mathbb{X}_i$ as $S_i=0$\;}
	 	\Else(\tcc*[f]{more than one layout has this \textbf{type set}}){
	 	\For{each pair $(x_j^i,x_k^i) (j \neq k)$ in $\mathbb{X}_i$}{ 
	 	calculate the IoU of bounding boxes between $x_j^i$ and $x_k^i$, denoted as $U_{jk}^i$\; }
average the $U_{jk}^i$ of the total ${l_i \choose 2}$ pairs to get the mean $S_i$\;}
 	 }
 	 \KwRet{the weighted average of all subsets' Self-Similarity score $\dfrac{\sum_{i=1}^n l_i S_i}{\sum_{i=1}^n l_i}$\;}

 	 \end{algorithm}
 \DecMargin{1em} 
 \vspace{-5px}

\begin{figure}[t]
  \centering
  \begin{subfigure}{0.28\linewidth}
    \includegraphics[width=1\linewidth]{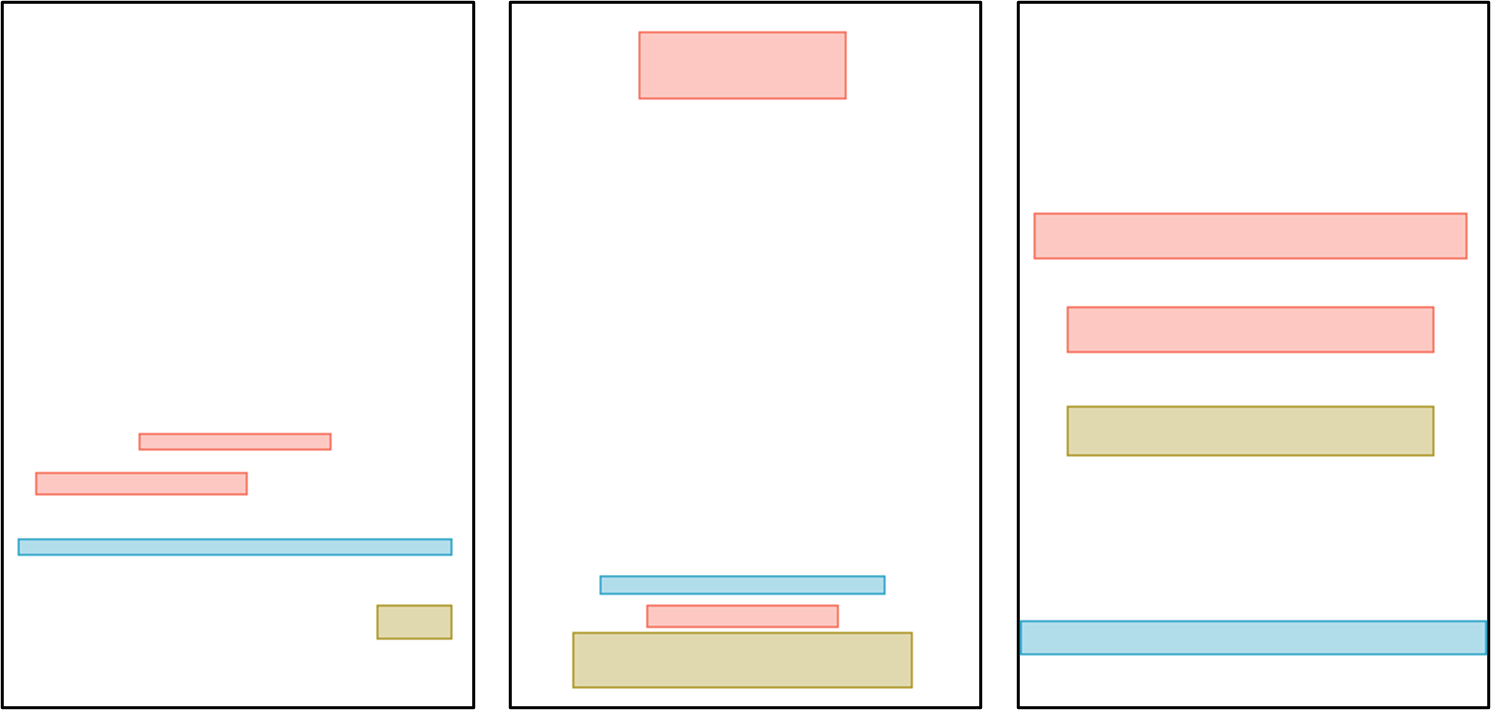}
    \caption{$S_i=0.0$}
    \label{fig:00}
  \end{subfigure}
  \hspace{10px}
  \begin{subfigure}{0.28\linewidth}
\includegraphics[width=1\linewidth]{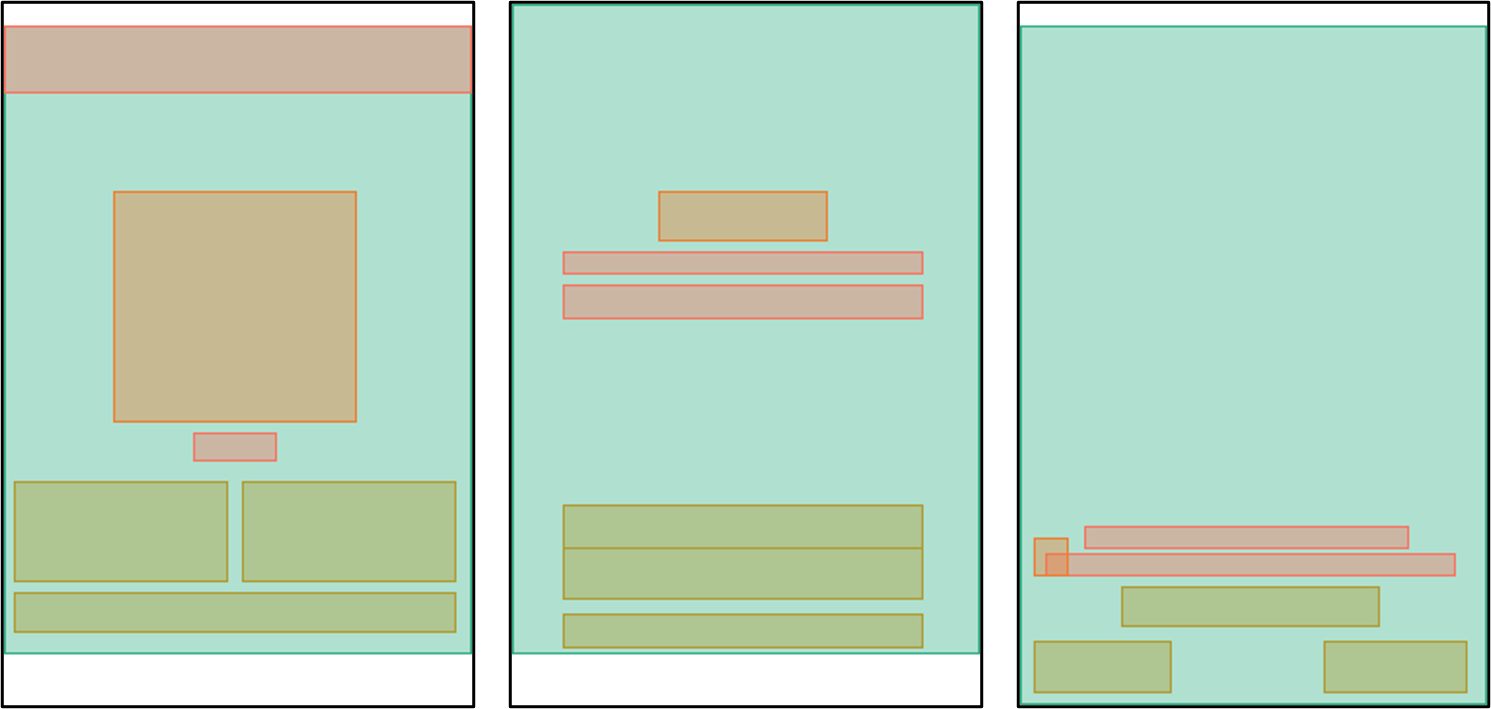}
    \caption{$S_i=0.2$}
    \label{fig:02}
  \end{subfigure}
  \hspace{7.5px}
    \begin{subfigure}{0.38\linewidth}
    \includegraphics[width=1\linewidth]{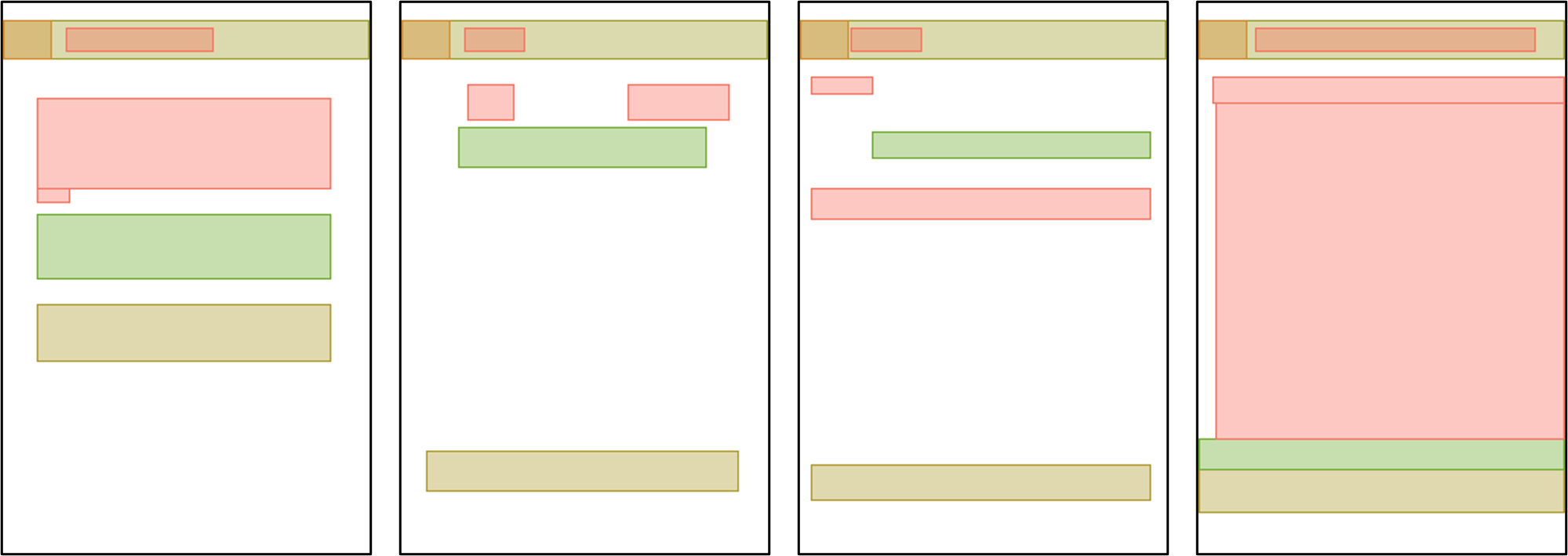}
    \caption{$S_i=0.4$}
    \label{fig:04}
  \end{subfigure}
    \vspace{5px}
    
  \begin{subfigure}{0.29\linewidth}
    \includegraphics[width=1\linewidth]{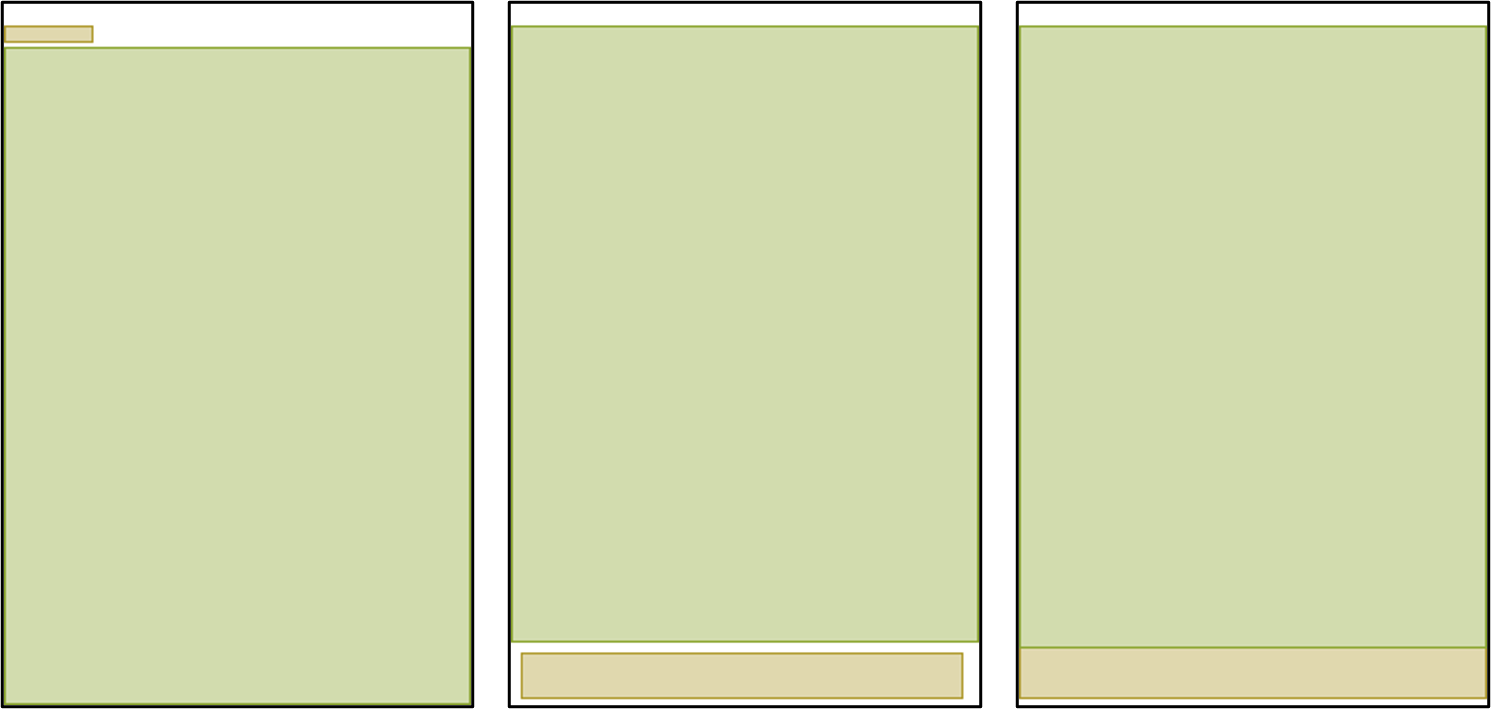}
    \caption{$S_i=0.6$}
    \label{fig:06}
  \end{subfigure}
  \hspace{10px}
  \begin{subfigure}{0.29\linewidth}
\includegraphics[width=1\linewidth]{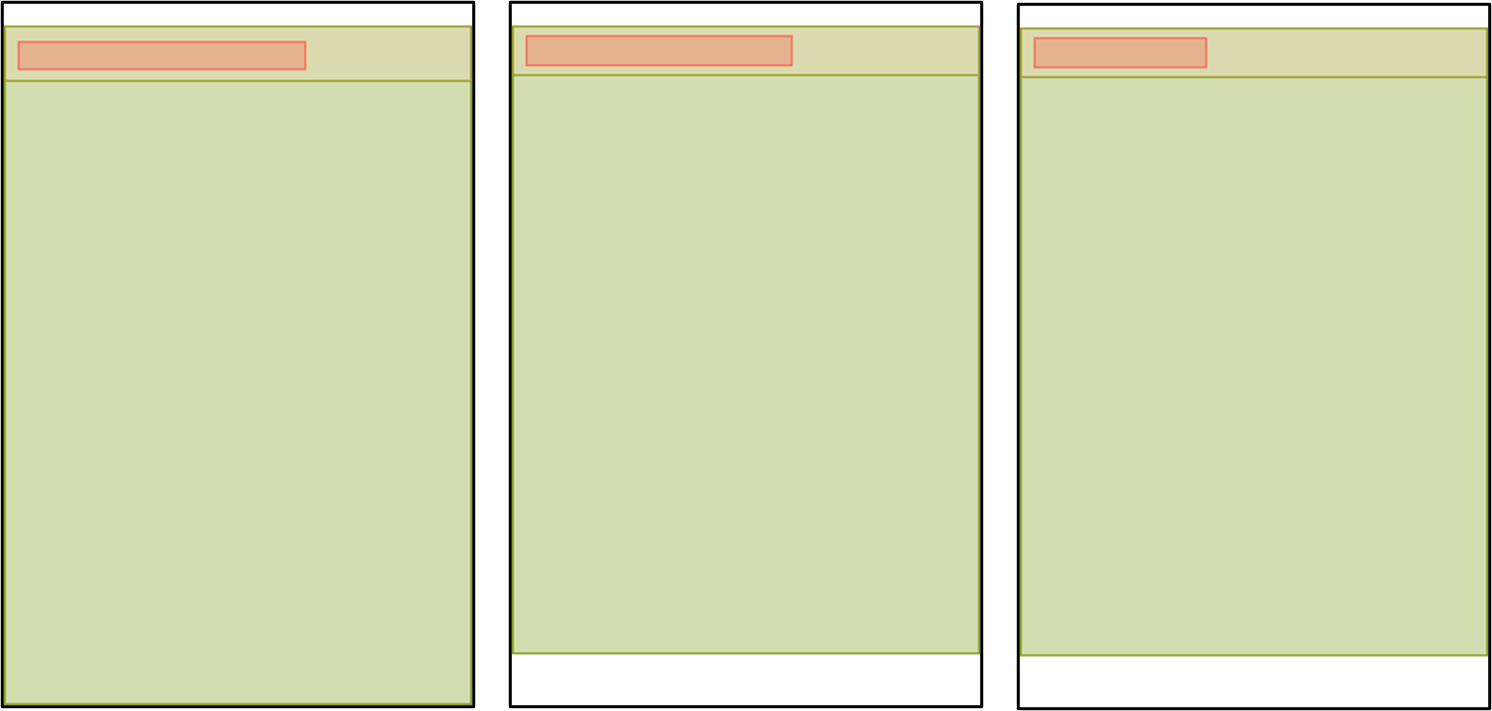}
    \caption{$S_i=0.8$}
    \label{fig:08}
  \end{subfigure}
  \hspace{10px}
    \begin{subfigure}{0.195\linewidth}
\includegraphics[width=1\linewidth]{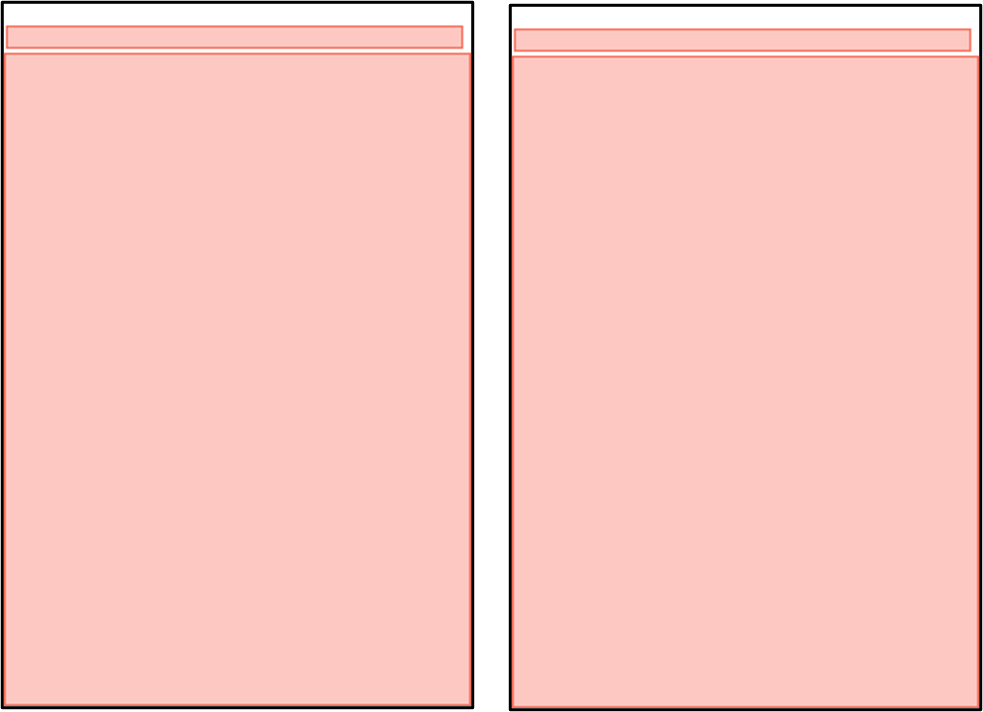}
    \caption{$S_i=1.0$}
    \label{fig:10}
  \end{subfigure}
  \caption{Examples of subsets with different SelfSim. As SelfSim goes from 0 to 1, the layouts in the subset go from totally different to completely identical.}
  \vspace{-5px}
  \label{fig:selfsim}
\end{figure}

\subsection{Case Study of SelfSim}
To visually demonstrate the effectiveness of SelfSim, we show some subsets with different SelfSims (i.e., subsets with different $S_i$) in~\cref{fig:selfsim}. We observe that subsets with higher SelfSims tend to have more similar layouts, while those with lower SelfSims have more diverse layouts. This further supports the effectiveness of the SelfSim metric for assessing the diversity of generated layouts.

\subsection{SelfSim Comparison with Existing Methods}
\cref{tbl:self-sim} compares LayoutDiffusion with existing layout methods and the diffusion-based method using SelfSim. While LayoutFormer++, Diffusion-LM, and LayoutTransformer have advantages in certain aspects of quality (as shown in \cref{Tab:quantitative_results} in the main paper), they suffer from obvious diversity issues, which aligns with our user study findings (as shown in \cref{user_study}). On the other hand, although D3PM performs slightly better in diversity on the PubLayNet dataset, it lags behind in terms of quality (see \cref{Tab:quantitative_results}). These results suggest that our proposed method achieves a better quality-diversity trade-off.

\begin{table}[t]
\centering
\renewcommand{\arraystretch}{0.9}
\setlength{\tabcolsep}{4pt}
\begin{tabular}{lccccccc}
\toprule

SelfSim$\downarrow$ & & LayoutTransformer & LayoutFormer++ & Diffusion-LM & D3PM (absorbing) & D3PM (uniform) & LayoutDiffusion\\
\midrule
RICO & & 0.318 & \textit{0.581} & \textit{0.326} & 0.157 & 0.165 & 0.157 \\
PublayNet & & \textit{0.314} & \textit{0.328} & 0.222 & 0.194 & 0.189 & 0.198 \\
\bottomrule
\end{tabular}
\vspace{-5px}
\caption{Comparison of SelfSim scores for unconditional generation on RICO and PublayNet datasets. Lower SelfSim scores indicate better diversity. \textit{Italic font} denotes the two worst-performing methods.}
\label{tbl:self-sim}
\end{table}

\section{Additional Experiments on Conditional Generation}
\label{more cond}
In this section, we design two sets of experiments to investigate LayoutDiffusion's robustness in the refinement task and its diversity performance in the generation conditioned on type (Gen-Type) task.

\subsection{Refinement}
\label{sec:suppl_refine}

\begin{table}[h]
    \centering
    \setlength{\tabcolsep}{7.5mm}{
    \renewcommand{\arraystretch}{1}
    \begin{small}
        \resizebox{\textwidth}{!}{
            \begin{tabular}{llcccc}
                \specialrule{1.1pt}{0pt}{1pt}

Noise level & \makecell[c]{Methods}       & mIoU $\uparrow$  &   Overlap$\rightarrow$ & Align.  $\rightarrow$  & FID  $\downarrow$  \\ \midrule
\multirow{3}{*}{std.=0.005} & RUITE                 & 0.743 & 0.473   & 0.126               & 1.244   \\
      & LayoutFormer++             & 0.722 & 0.479   & 0.119               & 1.043   \\
      & LayoutDiffusion   (30 steps)  & \textbf{0.787} & \textbf{0.467}   & \textbf{0.095}               & \textbf{0.499}   \\\midrule
\multirow{3}{*}{std.=0.01}  & RUITE                  & 0.716 & 0.483   & 0.139               & 1.475   \\
      & LayoutFormer++             & 0.704 & 0.487   & 0.123               & 1.124   \\
      & LayoutDiffusion   (40 steps)  & \textbf{0.759} & \textbf{0.467}   & \textbf{0.098}               & \textbf{0.500}   \\\midrule
\multirow{3}{*}{std.=0.02}  & RUITE                  & 0.611 & 0.507   & 0.203               & 13.633  \\
      & LayoutFormer++              & 0.621 & 0.514   & 0.157               & 4.981   \\
      & LayoutDiffusion   (50 steps)  & \textbf{0.748} & \textbf{0.469}   & \textbf{0.097}               & \textbf{0.496}   \\\midrule
      & Real Data  &-& 0.466 & 0.093 & -       \\
                \specialrule{1.1pt}{1pt}{0pt}
            \end{tabular}}
    \end{small}}

        \caption{Qualitative comparison under different noise levels on RICO. The content in the brackets denotes for the number of inference steps. The best results are \textbf{bold}.}
    \label{Tab:refine_table}
\vspace{-5px}
\end{table}
\vspace{-5px}

As introduced in~\cref{sec:details cond tasks}, in the main paper, our experiments for the refinement task apply a mixture of different levels of noise as input. 
We suppose that the excellent results achieved by LayoutDiffusion are due to its capability of handling various levels of noise.
To investigate the model's robustness to the noise, in this section, we compare with the two strongest baselines and further study the performance of the methods under each specific noise levels.

Specifically, we evaluate the performance of different methods under the conditions that the standard deviation of the noise is 0.005, 0.01, and 0.02, respectively. Plus, for a fair comparison, we train the baselines with the input noise of 0.01 standard deviation, and apply the same model for inference.

\noindent \textbf{Quantitative results.} As shown in~\cref{Tab:refine_table}, for the two baselines (RUITE~\cite{rahman2021ruite} and LayoutFormer++~\cite{unilayout}), the models exhibit favorable performances when dealing with noise levels less than or equal to the training level (std.=0.005 and std.=0.01).
However, when the testing noise level is greater than the training's (std.=0.02), the models suffer a significant performance drop.
For LayoutDiffusion, it not only surpasses the baseline in all 12 competitions (3 levels $\times$ 4 metrics), but also consistently presents excellent performance as the noise level varies, indicating that LayoutDiffusion is highly robust to noise levels.

\begin{figure}[th]
  \centering

   \includegraphics[width=1.01\linewidth]{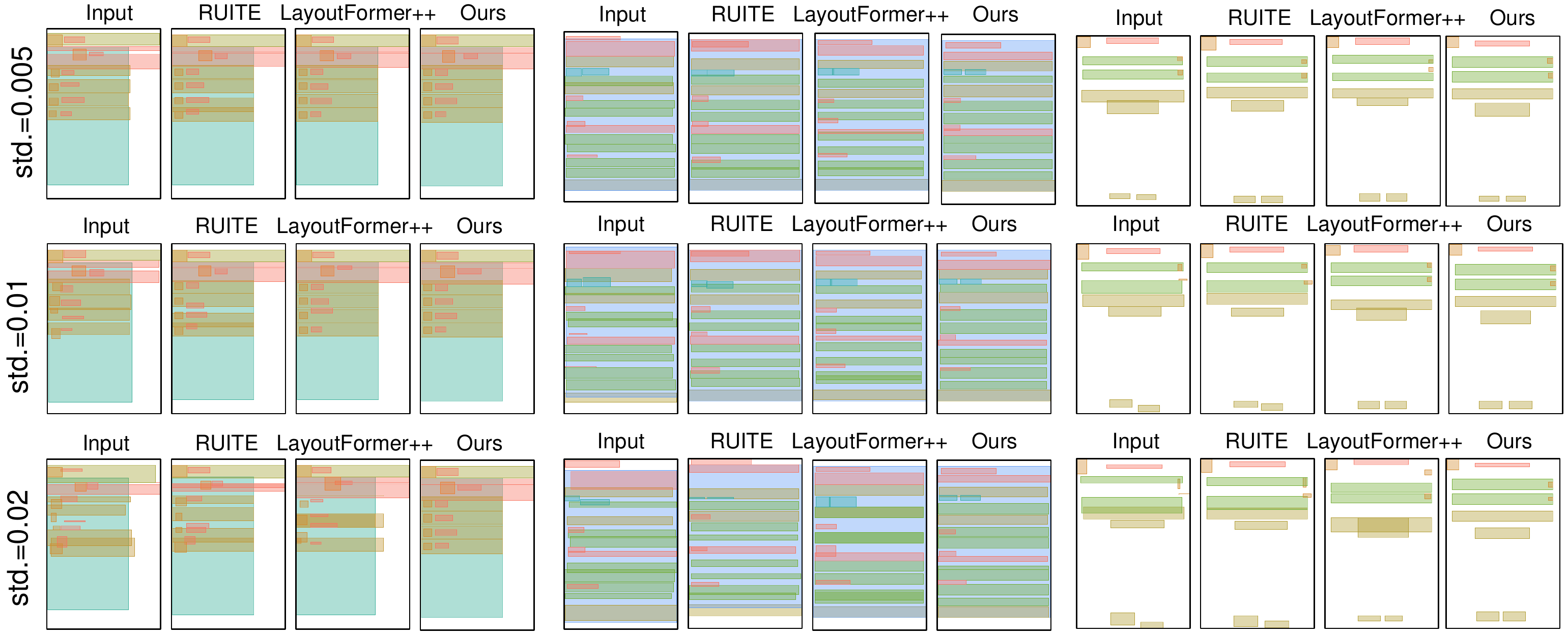}

   \caption{Qualitative comparison under different noise levels on RICO. Each row shares the same noise levels while each column shares the same method. For more quantitative result of LayoutDiffusion on Refinement, please refer to~\cref{more refine}.}
   \label{suppl refine} 
   \vspace{-5px}
\end{figure}
\noindent \textbf{Qualitative results.} We provide the quantitative results in~\cref{suppl refine}. We can conclude that as the input gets more chaotic, LayoutDiffusion consistently produces pleasing layouts, while other baselines fail to achieve so, which is in line with the quantitative results.

\subsection{Generation Conditioned on Type}
As described in~\cref{sec:details cond tasks}, in the main paper, our experiments for the Gen-Type task only generate one sample for each input type set.
In this section, we study whether they can generate multiple diverse layouts for a given type set to further explore the diversity performance of each method under Gen-Type task.

Specifically, we first find all the different type sets in the layouts of the test set, and then equally generate 5 layouts for each given type set\footnote{In practice, we find 2714 different type sets out of 3729 layouts in the test set of RICO, and 1339 type sets out of 10998 layouts in PublayNet's test set.}. 
We compare to the baseline with the best quality performance (i.e., LayoutFormer++~\cite{unilayout}), which can generate multiple layouts with top-k sampling~\cite{fan-etal-2018-hierarchical}. LayoutDiffusion can generate multiple layouts by simply running the inference process multiple times.
We apply the metric SelfSim (as discussed in~\cref{sec:selfsim}) for the evaluation of diversity.

\begin{table}[H]
    \centering
    \setlength{\tabcolsep}{6mm}{
    \renewcommand{\arraystretch}{1}
    \begin{small}
        \resizebox{\textwidth}{!}{
            \begin{tabular}{llccccc}
                \specialrule{1.1pt}{0pt}{1pt}

Dataset & \makecell[c]{Methods}       & mIoU $\uparrow$  &   Overlap$\downarrow$ & Align.  $\downarrow$  & FID  $\downarrow$ &SelfSim  $\downarrow$  \\ \midrule
\multirow{2}{*}{RICO}
      & LayoutFormer++           & \textbf{{0.375}}
&
0.563
&
0.125
&
9.786
&
0.536
  \\
      & LayoutDiffusion    & 0.357
&
\textbf{{0.490}}
&
\textbf{{0.062}}
&
\textbf{8.973}
&
\textbf{{0.268}}
 \\\midrule
\multirow{2}{*}{PublayNet} &  LayoutFormer++           & \textbf{{0.315}}
&
0.025
&
0.030
&
31.121
&
0.224
   \\
      & LayoutDiffusion     &0.312
&
\textbf{{0.007}}
&
\textbf{{0.029}}
&
\textbf{{21.522}}
&
\textbf{{0.189}}
   \\
                \specialrule{1.1pt}{1pt}{0pt}
            \end{tabular}}
    \end{small}}

        \caption{Qualitative comparison under new sampling strategy (5 samples for each type set). Since the type distribution of the generated layouts differs from that of the test set, we simply assume here that the less misalignment and overlap is better.}
    \label{Tab:type_table}
\vspace{-5px}
\end{table}

\noindent \textbf{Quantitative results.} The quantitative results is given in~\cref{Tab:type_table}. Compared to LayoutFormer++, LayoutDiffusion performs significantly better in diversity (as suggested by SelfSim), while achieving comparable quality performance (as suggested by Overlap and Align.).
We hypothesize that the gap between diversity is due to the probability accumulation of the autoregressive model while LayoutDiffusion samples each layout from independent noise.

\begin{figure}[th]
  \centering
  \vspace{-10px}
   \includegraphics[width=0.95\linewidth]{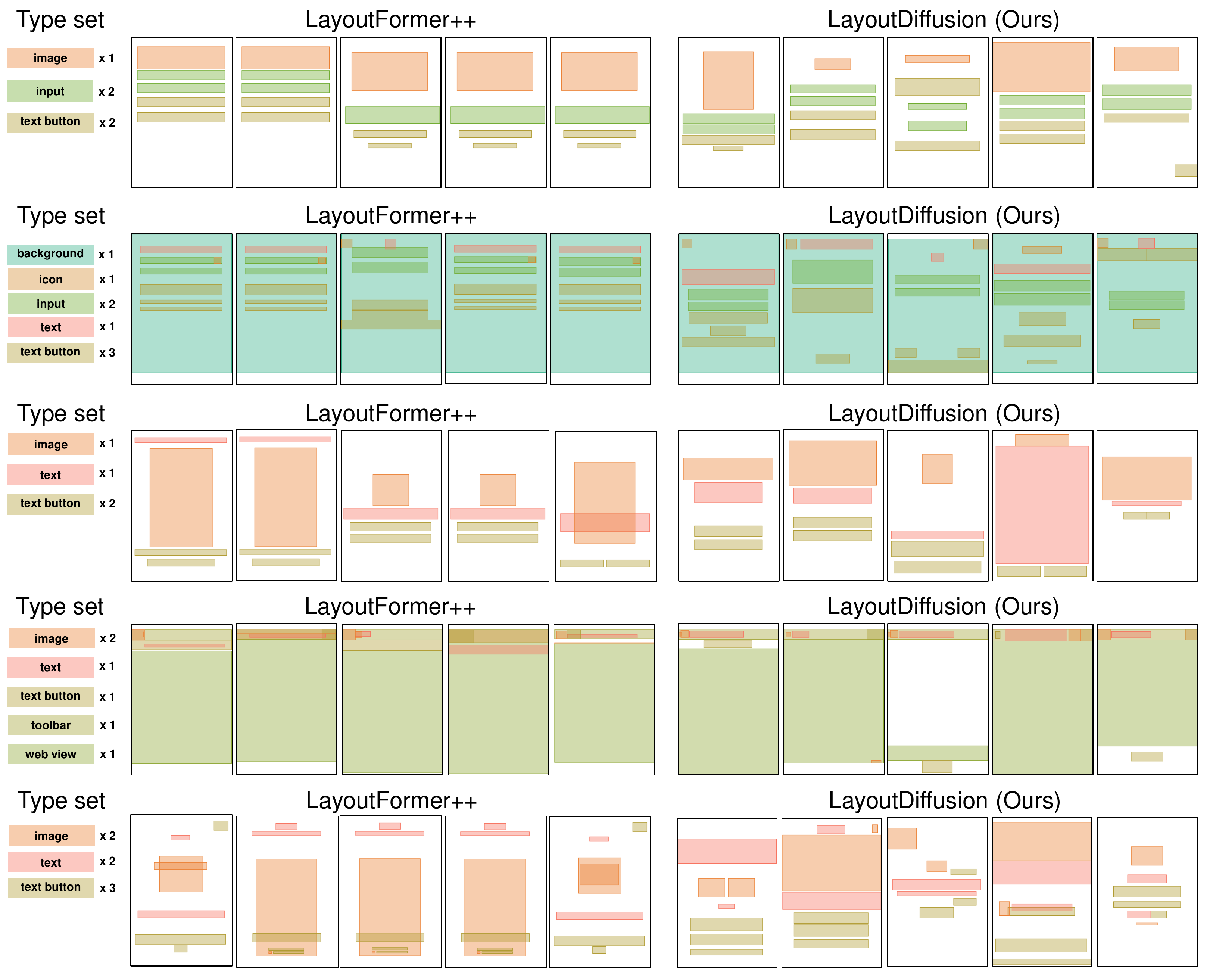}

   \caption{Qualitative comparison of diversity on RICO. Each type set corresponds to five samples from LayoutFormer++ (left) and five samples from LayoutDiffusion (right). For more qualitative results of LayoutDiffusion on Gen-Type, please refer to~\cref{more gen-type}.}
   \label{type suppl} 
\end{figure}

\noindent \textbf{Qualitative results.} As show in~\cref{type suppl}, despite equipped with top-k sampling, LayoutFormer++ still suffers severe diversity problem (duplication occurs in the first three row and the last row. Besides, the generated layouts in the fourth row share similar patterns). 
While for LayoutDiffusion, all the 5 samples of each type set are both pleasing and in great diversity, further demonstrating the superiority of LayoutDiffusion on Gen-Type task.

\section{Additional Ablation Studies}
\label{more ablation}
\subsection{Additional Ablation Studies on Conditional Generation Tasks}
In addition to the ablation studies on unconditional layout generation discussed in the main paper (see \cref{ablation}), we also conducted ablation studies with the variations on two conditional generation tasks, i.e., Gen-Type and Refinement, to further investigate the effectiveness of our design.
The quantitative comparison is shown in \cref{Tab:ablation_condition}. 

LayoutDiffusion consistently outperforms all the variations on almost all metrics in both tasks. This result indicates that our design is superior and the reverse generation process is well-suited to both tasks, allowing the model to better leverage the given conditions.
Notably, the comparison between LayoutDiffusion and Uniform $\mathbf{Q}_t^{\text{type}}$ as well as Linear $\overline{\gamma}_t$ highlights the importance of our handling of the type tokens, which considers type corruption factor. This factor leads to better utilization of type information in the Gen-Type task.
Moreover, the comparison between LayoutDiffusion and the two variations of $\mathbf{Q}_t^{\text{coord}}$ as well as Linear $\beta_t$ in the Refinement task demonstrates the importance of our design for the coordinate tokens, which helps us model the precise details of the layout and achieve better performance in the Refinement task.

\begin{table}[H]
    \centering
    \setlength{\tabcolsep}{3.5mm}{
    \renewcommand{\arraystretch}{0.75}
    \begin{small}
        \resizebox{\textwidth}{!}{
            \begin{tabular}{lcccccccc}
                \specialrule{1.1pt}{0pt}{1pt}
                                                                   & \multicolumn{4}{c}{Gen-Type}   & \multicolumn{4}{c}{Refinement}                                                                                                                                                                                 \\ \cmidrule(l){2-5} \cmidrule(l){6-9}
                 \makecell[c]{Methods} & mIoU $\uparrow$           &   Overlap $\downarrow$             & Align. $\downarrow$        & FID $\downarrow$& mIoU $\uparrow$          &   Overlap $\downarrow$          & Align.$\downarrow$       & FID$\downarrow$      \\ \midrule

                                             Uniform $\mathbf{Q}_t^{\text{coord}}$     &0.324 & 0.601 & 0.141 & 2.944   & 0.645 & 0.487 & 0.199 & 4.312      \\
                                             Absorbing $\mathbf{Q}_t^{\text{coord}}$                   &0.336 & 0.587 & 0.137 & 2.846     & - & -  & - & -     \\ 
                                            Uniform $\mathbf{Q}_t^{\text{type}}$          & 0.320& 0.532 &  0.188& 3.070 &0.698 & 0.477 &  0.167 & 2.443       \\ \midrule
                                            Linear $\overline{\gamma}_t$   & 0.308 & 0.513 & 0.164 & 2.768  & 0.667 & \textbf{0.467} & 0.133 &1.451 \\
                                            Linear $\beta_t$ &0.317 & 0.527 & 0.191 & 2.273  &0.659 & 0.491 & 0.185 &  1.835 \\\midrule
                                            LayoutDiffusion (ours)   &\textbf{0.345} & \textbf{0.491} & \textbf{0.124} & \textbf{1.557}   & \textbf{0.719} & 0.469 & \textbf{0.102} & \textbf{0.549}  \\
                                             
                \specialrule{1.1pt}{1pt}{0pt}
            \end{tabular}}
    \end{small}}
     \vspace{-5px}
        \caption{Quantitative results on conditional generation tasks for LayoutDiffusion and its ablations on RICO. The variation of absorbing $\mathbf{Q}_t^{\text{coord}}$ do not support refinement, as the coordinates are fixed during generation. The best result is in \textbf{bold}.
        }
    \label{Tab:ablation_condition}

\end{table}

\subsection{Additional Ablation Studies on Noise Schedule of Type Tokens}
In this section, we further investigate the effectiveness of our type schedule by experimenting other different $\overline{\gamma}_t$ schedules.

Recall that in the main paper, we follow the insight that type changes in the early stage may bring large semantic shift to the layout, thus, we set the noise schedule for $\overline{\gamma}_t$ as:
\begin{equation}
    \overline{\gamma}_t = \begin{cases}
    0, &t< \Tilde{T} \\
(t-\Tilde{T})/(T-\Tilde{T}), &t \ge \Tilde{T} 
\label{late absorb}
\end{cases}
\end{equation}

We denote this kind of schedule as ``late absorb $\Tilde{T}$", since under this schedule, all type tokens stay unchanged until timestep $\Tilde{T}$ when they start to absorb, and at the terminal step $T$, all type tokens reach the absorbed state. Follow this idea, we can come to a similar noise schedule, ``early absorb $\Tilde{T}'$", where the type tokens start to absorb at the beginning and fully adsorbed in the early stage, and it can be defined as follows:

\begin{equation}
    \overline{\gamma}_t = \begin{cases}
    t/\Tilde{T}', &t< \Tilde{T}' \\
1, &t \ge \Tilde{T}' \\
\end{cases}
\end{equation}

Note that, when $\Tilde{T}'=T$ and $\Tilde{T}=0$, two schedules becomes the same and is experimented in the ablation studies of the main paper (denoted as ``linear $\overline{\gamma}_t$"). Here, we provide a detailed experiment on $\overline{\gamma}_t$, including different choices of ``late absorb $\Tilde{T}$" and ``early absorb $\Tilde{T}'$".

\begin{table}[h]
    \centering
    \setlength{\tabcolsep}{6.25mm}{
    \renewcommand{\arraystretch}{0.95}
    \begin{small}
        \resizebox{\textwidth}{!}{
            \begin{tabular}{llcccc}
                \specialrule{1.1pt}{0pt}{1pt}
Experiments & \makecell[c]{Methods}       & mIoU $\uparrow$  &   Overlap$\rightarrow$ & Align.$\rightarrow$  & FID  $\downarrow$  \\ \midrule
\multirow{4}{*}{Type schedule}  & early   absorb $ \Tilde{T}' = 40$                    & 0.574 & 0.512 & 0.160 & 3.107 \\
 & early   absorb $\Tilde{T}' = 100$                   & 0.590 & 0.496 & 0.143 & 2.952 \\
 & late absorb $\Tilde{T} = 0$ (early absorb $\Tilde{T}' = 200$)   & 0.580 & 0.522 & 0.156 & 2.846 \\
 & late absorb $\Tilde{T} = 100$                    & 0.599 & 0.495 & 0.121 & 2.612 \\\midrule
 \multirow{5}{*}{Sequence ordering}& ltwh+random                          & 0.585 & 0.505 & 0.136 & 3.166 \\
 & ltwh+position                        & 0.577 & 0.491 & 0.128 & 3.055 \\
 
 & ltwh+lexico                          & 0.578 & 0.501 & 0.125 & 3.111 \\
 & ltrb+random                          & 0.619 & 0.504 & 0.089 & 2.505 \\
 & ltrb+position                        & 0.613 & 0.482 & 0.115 & 2.446 \\
 \midrule
 \multirow{1}{*}{Ours}&    ltrb+lexico, late absorb $\Tilde{T} = 160$ & 0.620 & 0.502 & 0.069 & 2.490\\\midrule
 Real Data  &&-& 0.466 & 0.093 & -      \\
                \specialrule{1.1pt}{1pt}{0pt}
            \end{tabular}}
    \end{small}}
        \caption{More ablation studies under unconditional generation task on RICO.}
    \label{Tab:ablation_table}
\vspace{-5px}
\end{table}

As shown in the first group of~\cref{Tab:ablation_table}, we can conclude that as the type starts absorb later (from top to bottom in the table), the overall generation performance becomes better. Specifically, with $\Tilde{T}$ for the late absorb decreases, the quality of the generated layouts gets worse (as suggested by mIoU, Align., and FID). When it comes to early absorb, the performance drops as $\Tilde{T}'$ decreases.
The experiment empirically supports the insight we discuss above.

\subsection{Ablation Studies on Sequence Ordering}
In the main paper, we sort the layout sequence according to the alphabetical order of the elements' type in the layout (denoted as ``lexico"). Other choices are the positional ordering of the elements' bounding boxes (denoted as ``position") or simply randomly sorting the elements (denoted as ``random"). 
Besides, for each element in the layout, we represent its bounding box by the left, top, right, bottom coordinates (denoted as ``ltrb'').
One can also represent the bounding box using an element's left coordinate, top coordinate, width and height (denoted as ``ltwh'').
Here, we provide the results of all these options for sequence ordering.

As shown in the second group of~\cref{Tab:ablation_table}, for the format representing the coordinates, the ltrb group exhibits better overall performance than the ltwh group. One explanation is that ltrb format may provide more straightforward information for the precise alignment of the bounding boxes. 
For the ordering of elements, the alphabetical ordering and positional ordering are slightly better than the random ordering. We hypothesize that the model can exploit the additional ordering information with positional embedding. 
However, note that for conditional generation, the positional ordering of the elements is unknown, so to enable both unconditional and conditional generation, we apply the alphabetical ordering to sort the elements.

\section{Qualitative Results of LayoutDiffusion}
\label{sec:suppl_quali}
In this section, we provide more generated samples covering three unconditional and conditional tasks on two datasets. 
\subsection{Unconditional Generation}
\begin{figure}[H]
  \centering
   \includegraphics[width=0.9\linewidth]{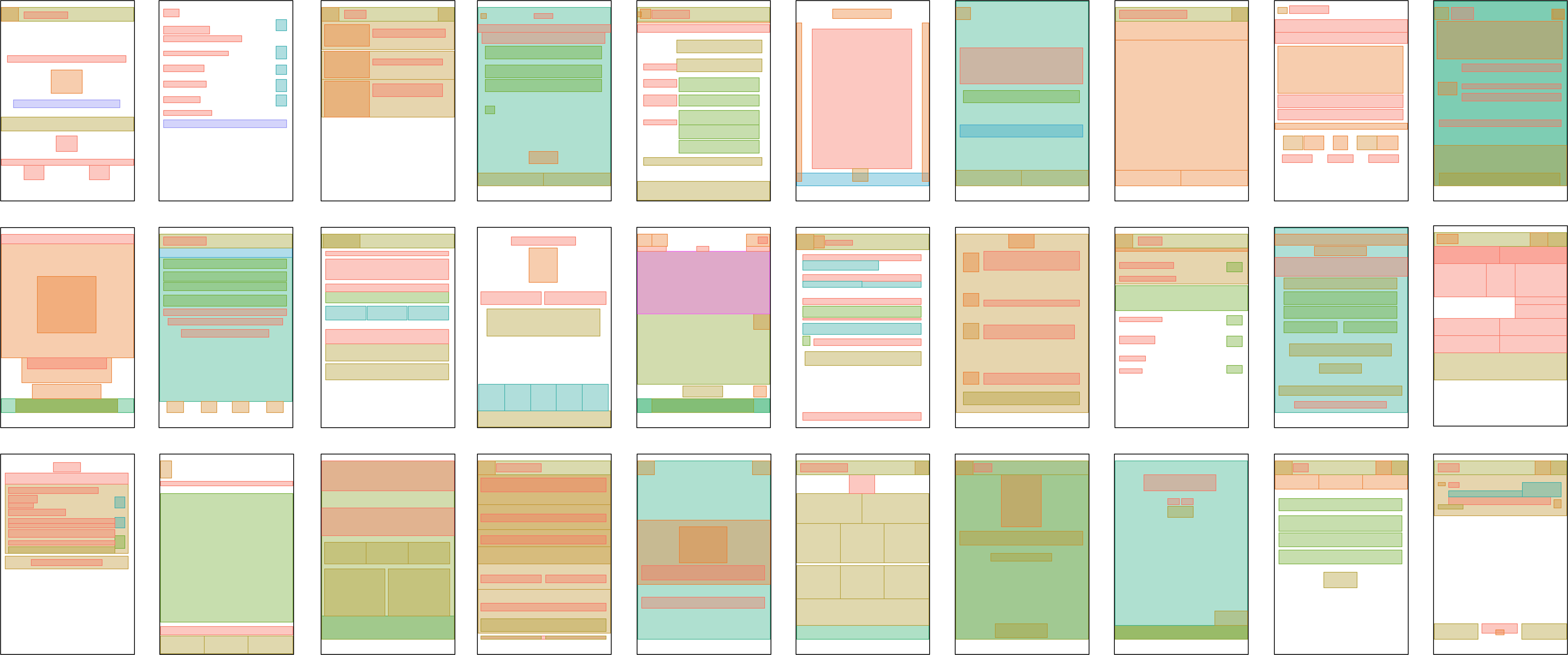}
   \caption{Examples of unconditional generation on RICO dataset.}
   \label{ungen_rico}
\end{figure}
\begin{figure}[H]
  \centering
   \includegraphics[width=0.9\linewidth]{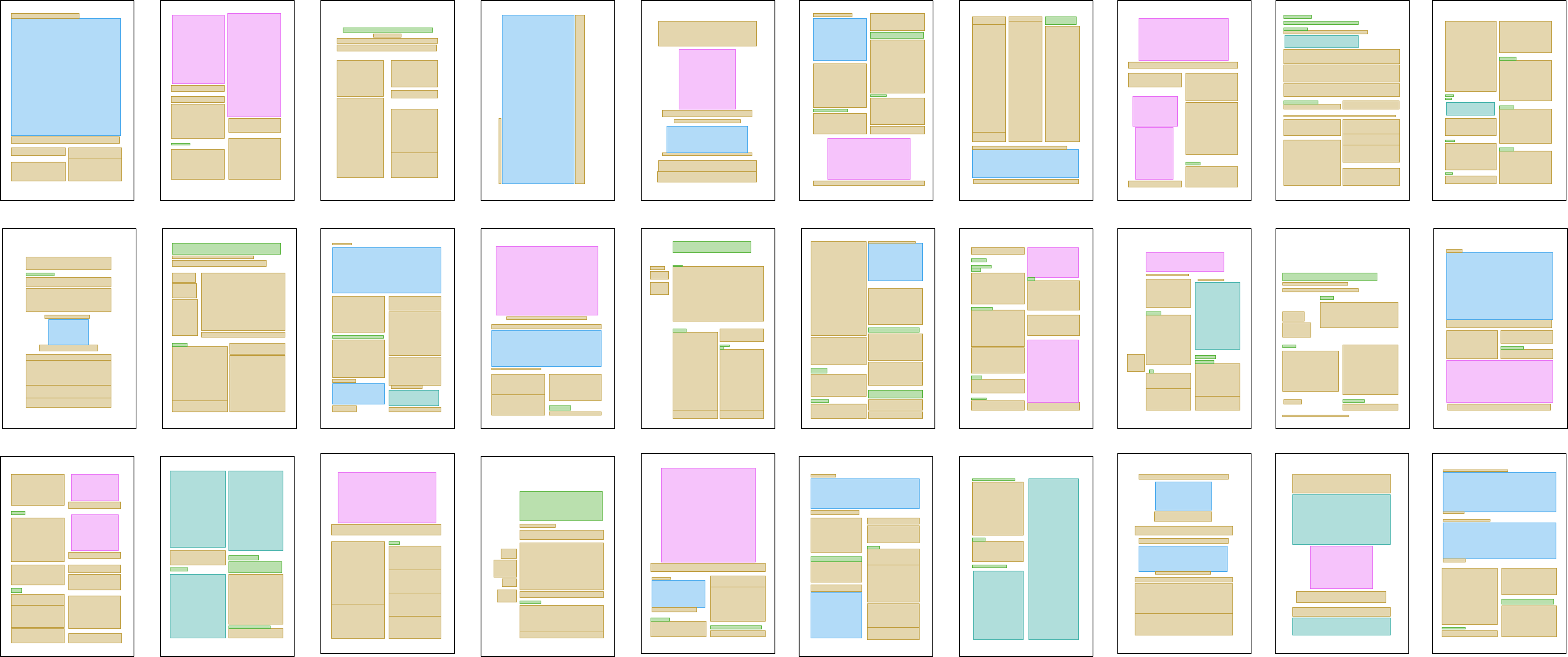}
   \caption{Examples of unconditional generation on PublayNet dataset.}
   \label{ungen_pub}
\end{figure}

\subsection{Refinement}

\label{more refine}
\begin{figure}[H]
  \centering
   \includegraphics[width=0.925\linewidth]{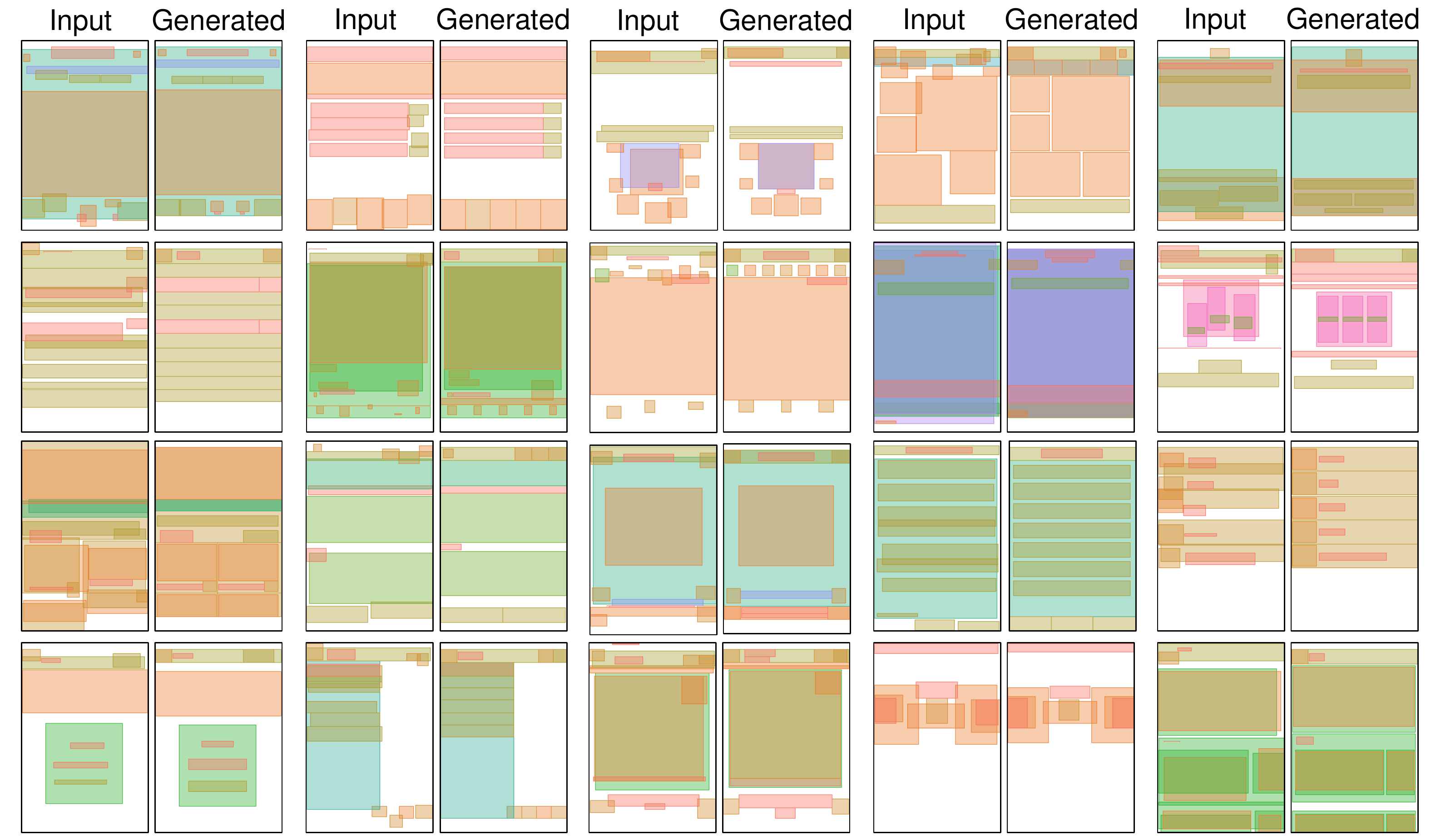}
\caption{Examples of refinement on RICO. The left side of each pair is the input layout while the right side is the generated layout.}
   \label{refine rico} 
\end{figure}

\begin{figure}[H]
  \centering
   \includegraphics[width=0.925\linewidth]{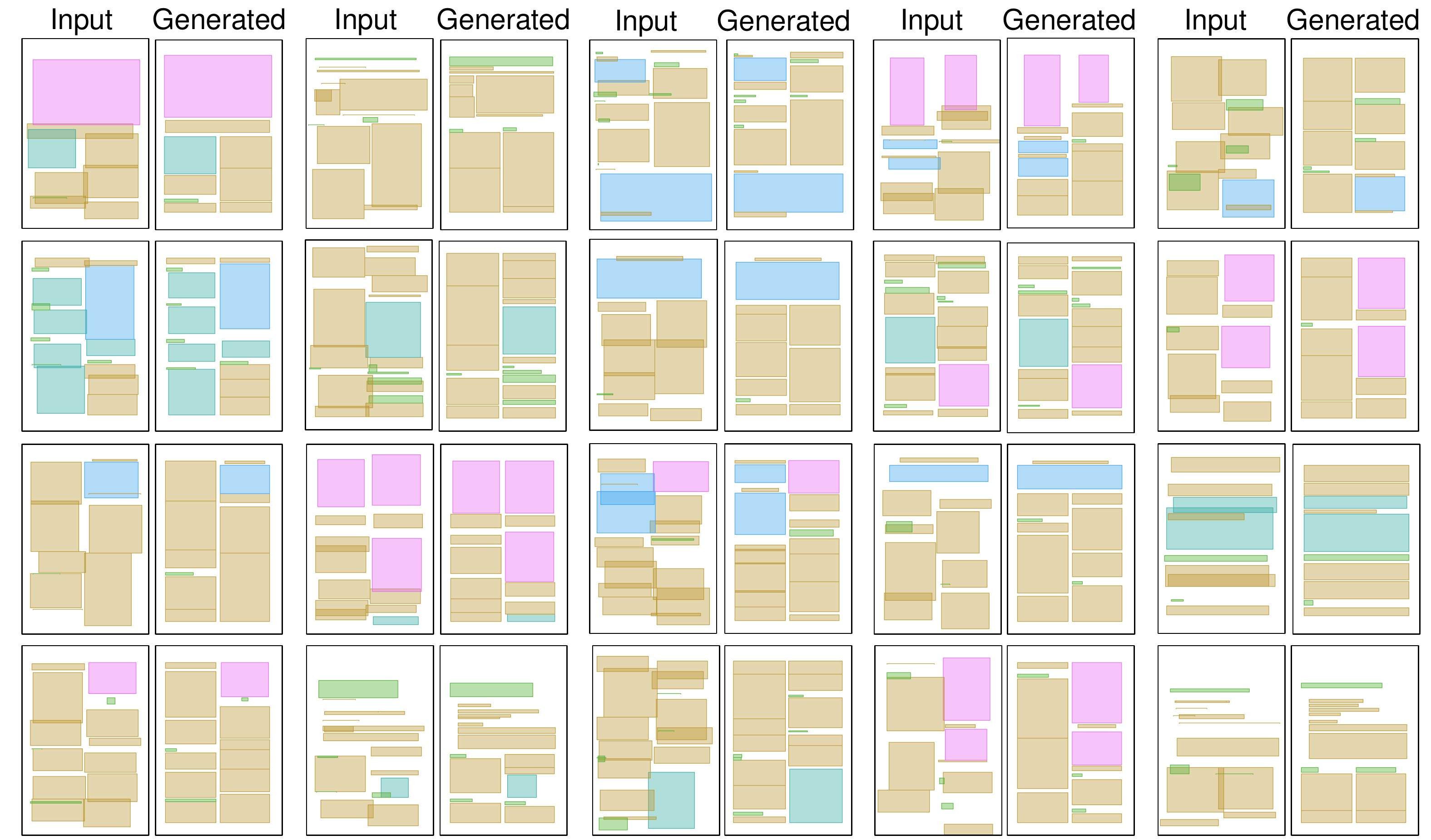}
\caption{Examples of refinement on PublayNet. The left side of each pair is the input layout while the right side is the generated layout.}
   \label{refine pub}
\end{figure}

\subsection{Generation Conditioned on Type}
\label{more gen-type}
\vspace{-10px}
\begin{figure}[H]
  \centering
   \includegraphics[width=0.9\linewidth]{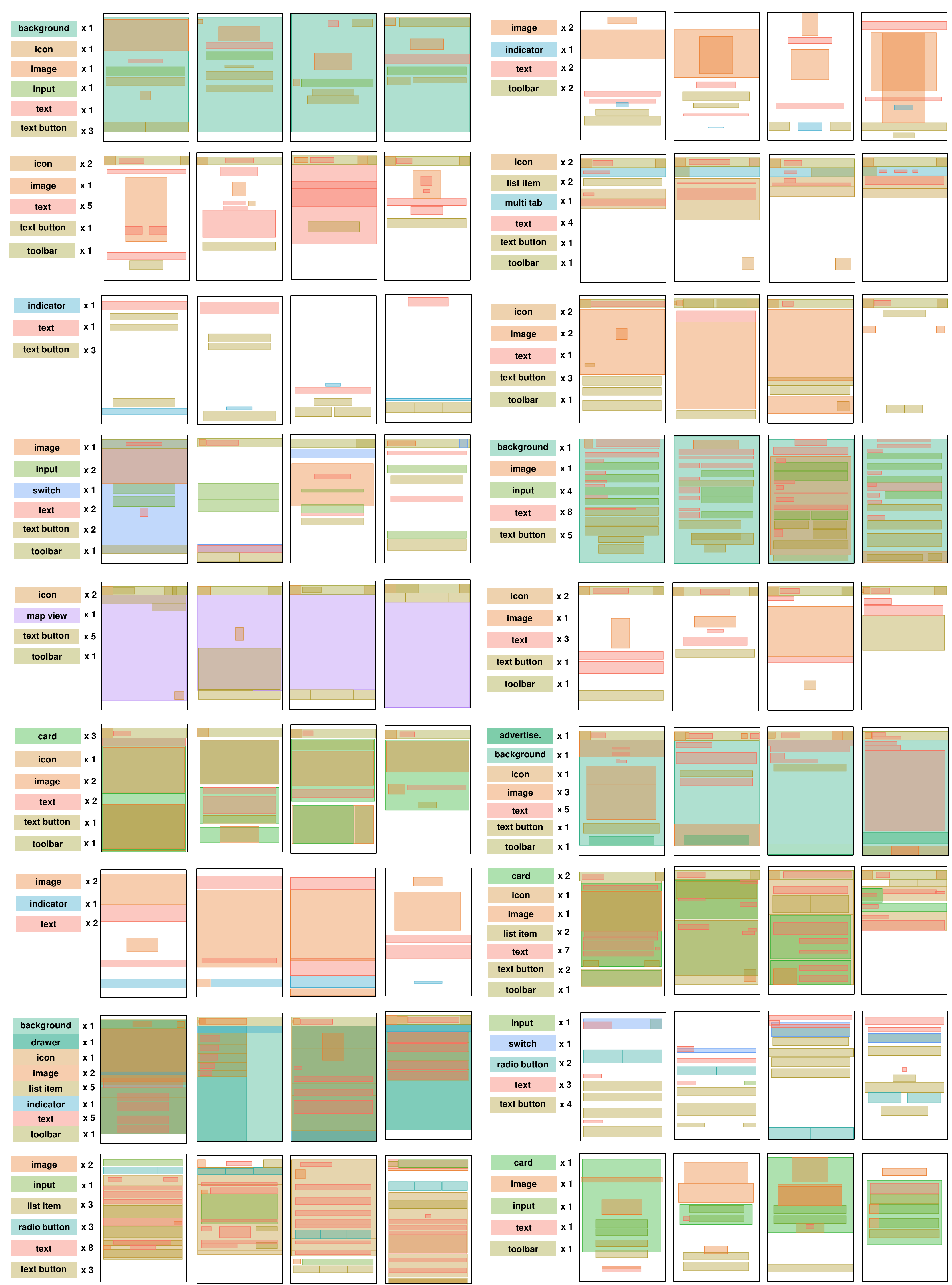}
\caption{Examples of Gen-Type on RICO. Each given type set corresponds to four generated layouts.}
   \label{type rico}
\vspace{-5px}
\end{figure}

\begin{figure}[H]
  \centering
   \includegraphics[width=0.93\linewidth]{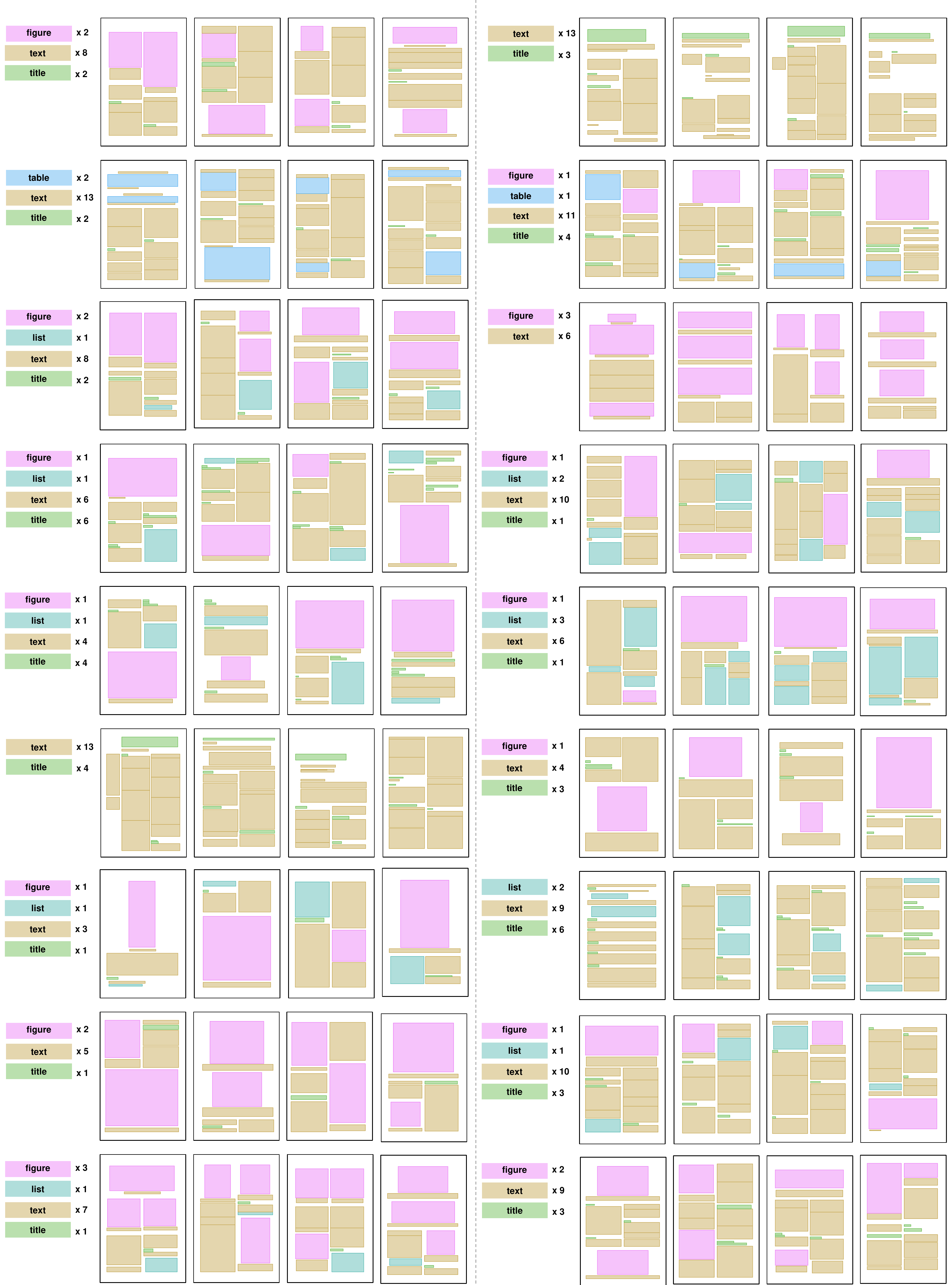}
   \vspace{-5px}
\caption{Examples of Gen-Type on PublayNet. Each given type set corresponds to four generated layouts.}
   \label{type pub} 
\end{figure}

\section{Fine-Grained Visualization of the Forward Diffusion Process}
\begin{figure}[H]
  \centering
   \includegraphics[width=0.925\linewidth]{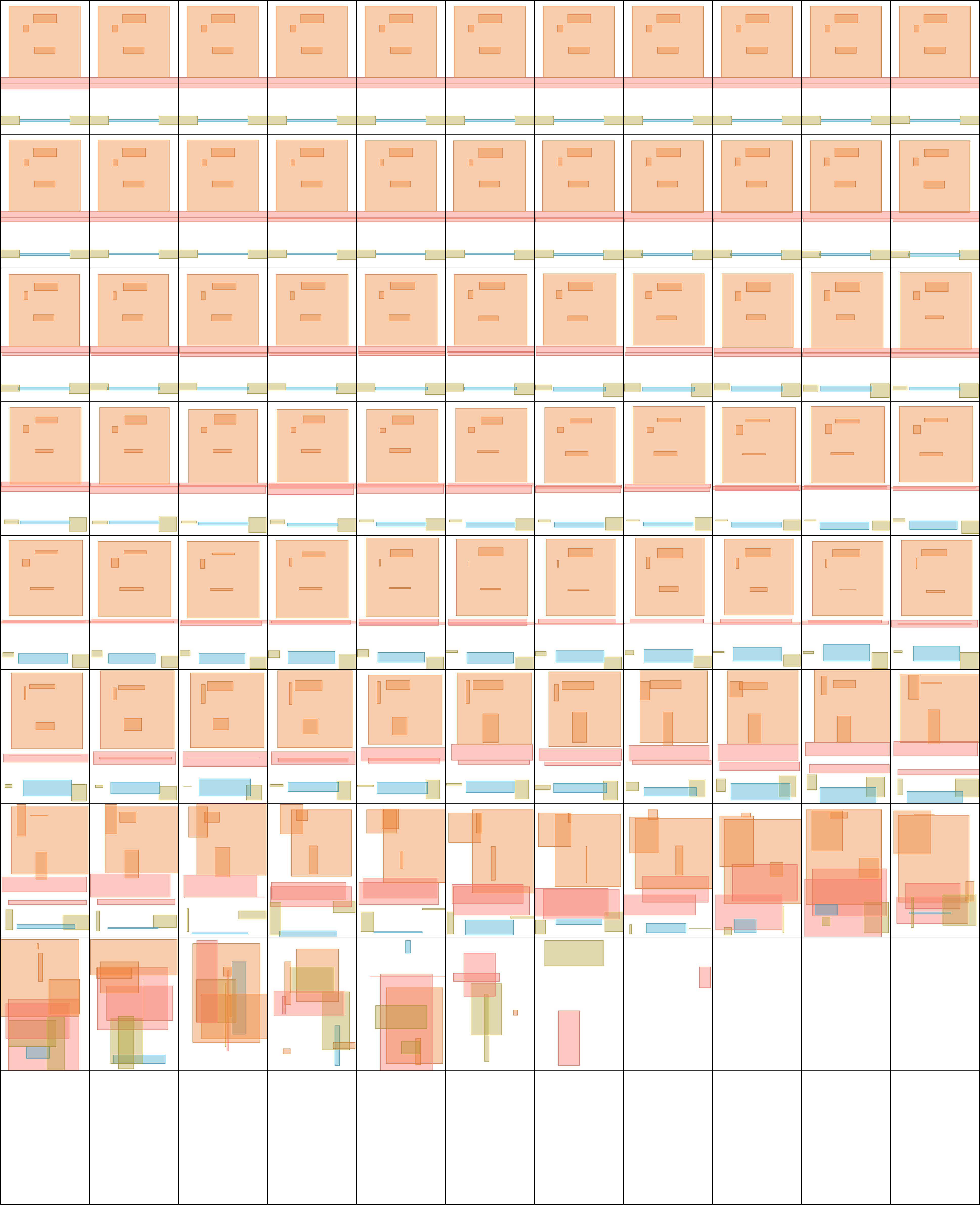}
   \caption{An example of the forward process of LayoutDiffusion on RICO. We sample 99 steps uniformly from the total 200 timesteps.}
   \label{noise steps} 
\end{figure}

\begin{figure}[H]
  \centering
   \includegraphics[width=1.01\linewidth]{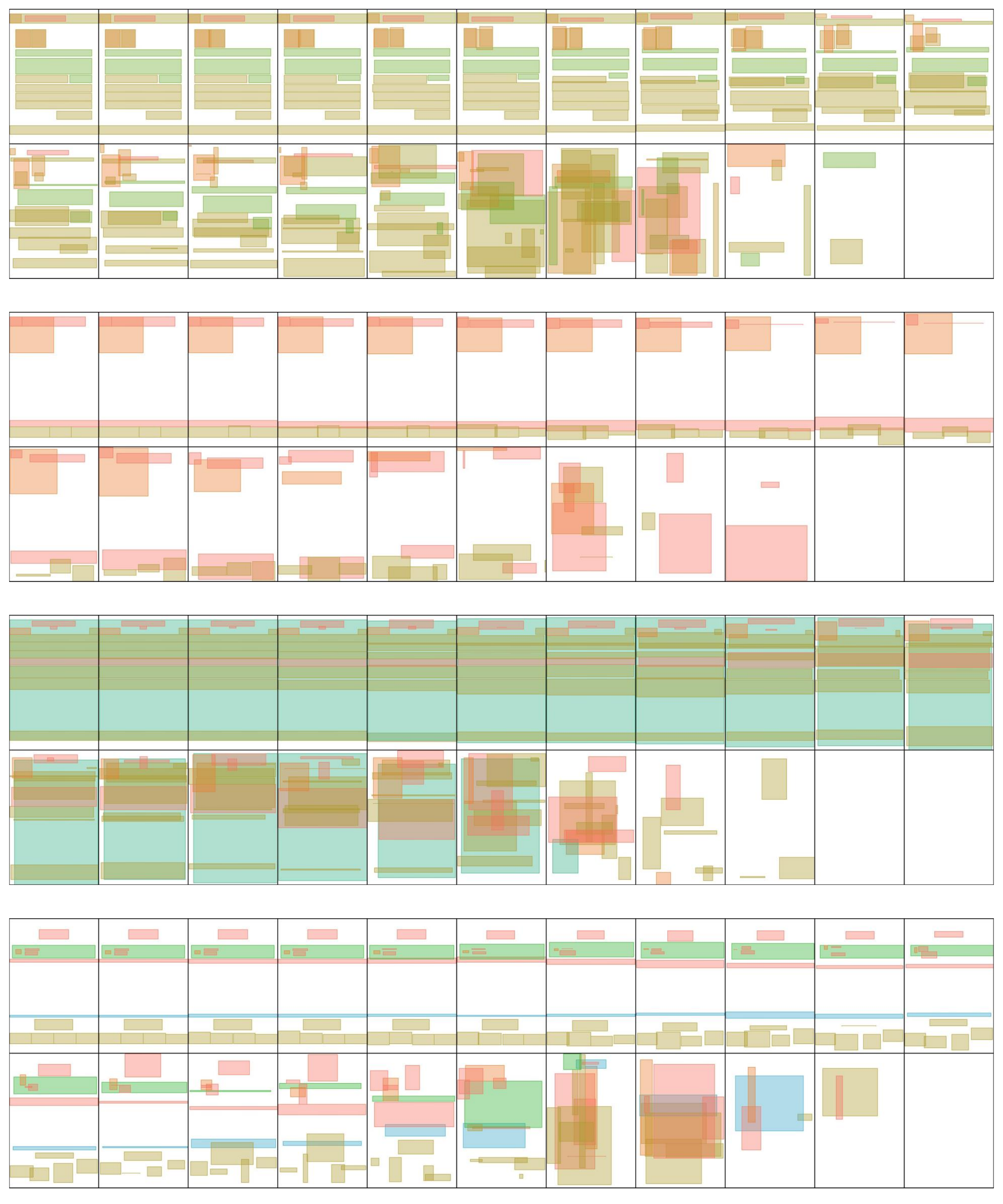}
   \caption{Examples of the forward process of LayoutDiffusion on RICO. We sample 22 steps uniformly from the total 200 timesteps.}
   \label{noise steps more} 
\end{figure}

\section{Fine-Grained Visualization of the Reverse Generation Process}
\begin{figure}[H]
  \centering
   \includegraphics[width=0.925\linewidth]{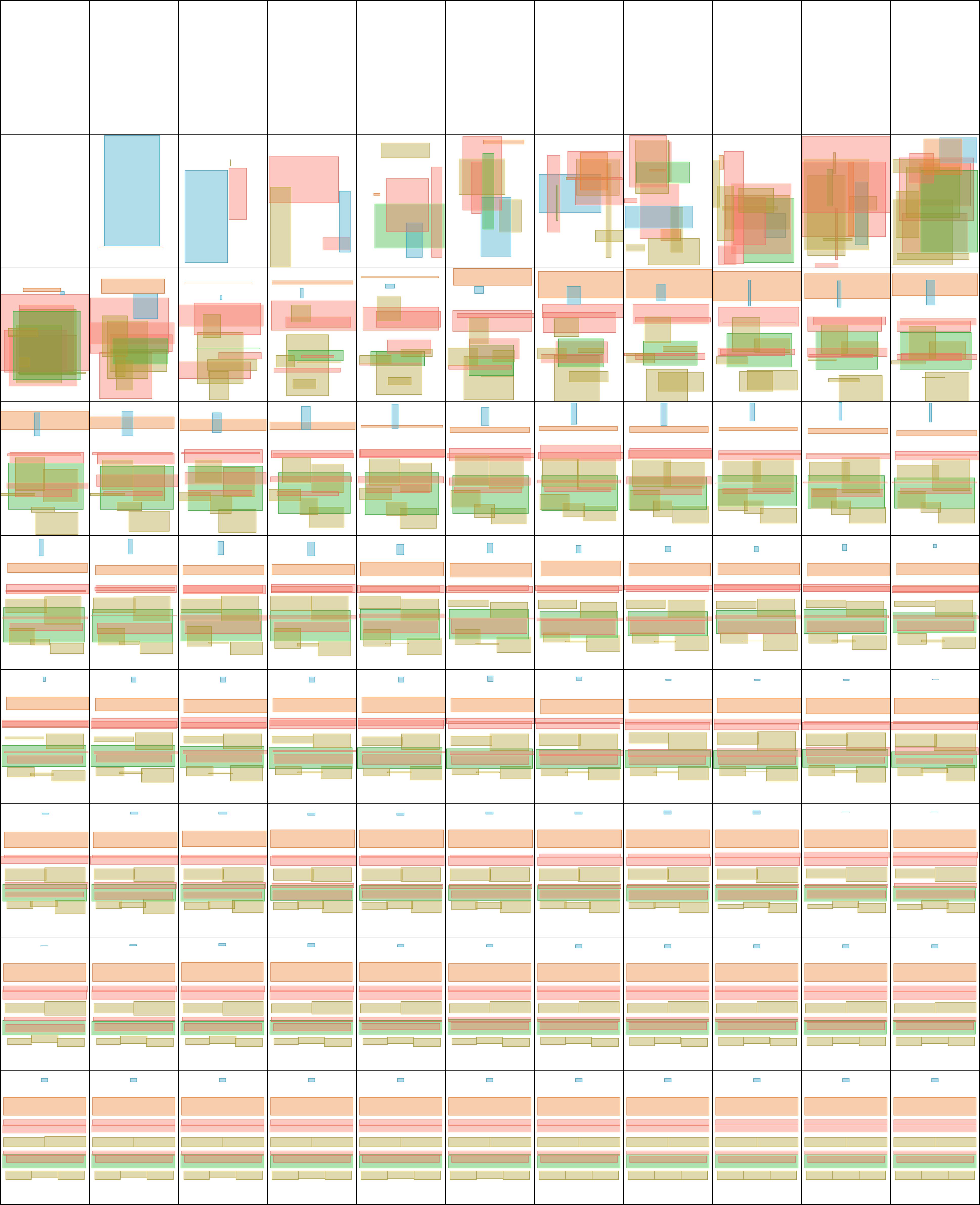}
   \caption{An example of the reverse process of LayoutDiffusion on RICO. We sample 99 steps uniformly from the total 200 timesteps.}
   \label{reverse steps} 
\end{figure}

\begin{figure}[H]
  \centering
   \includegraphics[width=1.01\linewidth]{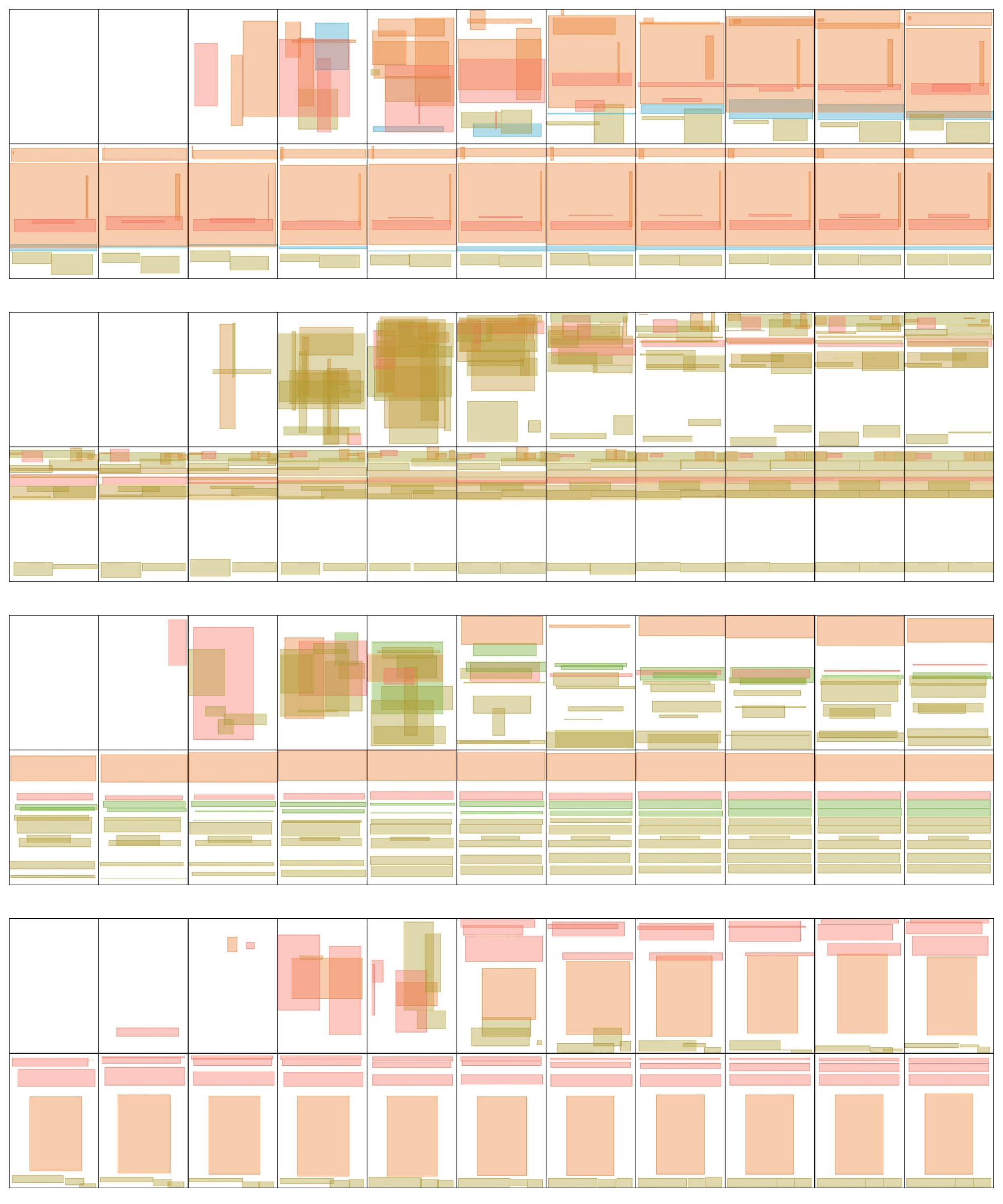}
   \caption{Examples of the reverse process of LayoutDiffusion on RICO. We sample 22 steps uniformly from the total 200 timesteps.}
   \label{reverse steps more} 
\end{figure}

\end{document}